\documentclass[3p,12p]{elsarticle}

\usepackage{amsmath}
\usepackage{amssymb}
\usepackage{amsfonts}
\usepackage{acronym}
\usepackage{amsthm}
\usepackage{stmaryrd}
\usepackage{xcolor}
\usepackage{soul}
\usepackage{booktabs}
\usepackage{graphicx}
\usepackage{todonotes}
\usepackage{subcaption}
\usepackage{tikz}
\usetikzlibrary{arrows.meta}

\newtheorem{example}{Example}

\DeclareMathOperator*{\maximize}{maximize}

\graphicspath{{figures/}}

\journal{Information Sciences}

\acrodef{MAE}{Mean Absolute Error}
\acrodef{DM}{Decision Maker}

\title{From Validator Selection to Portfolio Collection Optimization in~Proof-of-Stake Blockchains}

\author[1]{Jonas Gehrlein}
\ead{gehrlein.jonas@protonmail.com}
\author[put]{Grzegorz Miebs\corref{cor}}
\ead{grzegorz.miebs@cs.put.poznan.pl}
\author[3]{Matteo Brunelli}
\ead{matteo.brunelli@unitn.it}
\author[put]{Adam Mielniczuk}
\author[put]{Mi\l{}osz Kadzi\'nski}
\ead{milosz.kadzinski@put.poznan.pl}
\cortext[cor]{Corresponding author: tel. +48-61 665 3022; fax: +48-61 8771 525.}
\address[1]{Parity Technologies AG, Zug, Switzerland}
\address[put]{Institute of Computing Science, Poznan University of Technology, Pozna{\'n}, Poland}
\address[3]{Department of Industrial Engineering, University of Trento, Italy}

\begin{document}

\begin{abstract}
We consider a problem arising in proof-of-stake blockchain environments, where agents called nominators select validators---entities responsible for maintaining the blockchain's physical infrastructure. The selection process is inherently subjective and multi-criterial and combines with the fact that nominators commonly operate through multiple accounts. This gives rise to a portfolio selection problem, where agents seek to distribute their nominations across accounts to diversify risk. We propose a decision support framework to optimize this selection by simultaneously maximizing two objectives: the expected utility of the validators likely to be allocated, representing portfolio quality and profitability, and the expected entropy of the allocation, representing diversification and risk mitigation across stashes.
Validator utilities are derived using an original active preference learning procedure based on multi-attribute value theory, with emphasis on top-ranked validators.
The resulting bi-objective optimization problem is solved with a multi-objective evolutionary algorithm and, to support the final choice, we introduce an interactive binary search navigation procedure that guides the nominator through the front and identifies a satisfactory trade-off with only a few questions. Numerical experiments examine the optimization strategies, while an expert assessment involving five experienced nominators confirms the approach's practical relevance and usefulness.
\end{abstract}


\maketitle


\section{Introduction}
Blockchain technology has evolved into a foundational infrastructure for decentralized systems spanning finance, supply chains, governance, and the broader Web3 ecosystem \citep{HuangEtAl2024}. Its reach now extends well beyond cryptocurrency, with recent work leveraging distributed ledgers for secure healthcare data management and the realization of sustainable digital health ecosystems \citep{sadeghi2024synergy}, as well as for tamper-proof authentication protocols in domains as diverse as autonomous vehicle communication \citep{Kumar2026blockchain}. Across these settings, the value of blockchain rests on the integrity of its underlying consensus mechanism. Proof-of-stake (PoS) protocols address this challenge by requiring participants to lock financial resources (so-called \emph{stake}) as collateral, which is exposed to penalties if the actor is caught deviating from the protocol \citep{Saleh2021, BentovEtAl2014}. Honest behavior is thereby aligned with economic incentives. Yet, such incentives only safeguard the network as long as a sufficient supermajority of active validators is genuinely trustworthy, a condition rooted in the classical theory of Byzantine fault tolerance \citep{lamport2019byzantine}: if a critical fraction of validators colludes or misbehaves, the system's security guarantees collapse regardless of how the underlying consensus mechanism is designed. 

Selecting and maintaining this set of benevolent validators is therefore the central challenge in an open PoS network, and it is most frequently handled by a distinct class of actors: \emph{nominators} (often called delegators). Nominators are stakeholders who curate the pool of validator candidates and submit preference lists of those they deem trustworthy and capable. They must collectively elect a supermajority of honest validators, which are at the same time sufficiently capable of running and maintaining the network infrastructure. The evaluation of the fitness of a validator spans many relevant dimensions and cannot be reduced to a fixed rule, which is why this judgment is delegated to human stakeholders.


These lists are then aggregated by an election algorithm to determine the \emph{active set} of validators for each period. The system is financially incentivized: nominators share the rewards earned by their nominated validators, receiving higher payouts when validators perform well and lower payouts when they underperform. In addition, nominators bear the risk of slashing, a partial or total loss of staked funds, if their validators misbehave. This design ensures that the quality of the active validator set, and hence the security and reliability of the entire network, depend critically on nominators' decisions~\citep{GehrleinEtAl2023}.

In this study, we focus on the Polkadot network, one of the largest PoS blockchain platforms, consistently ranked among the highest cryptocurrencies by market capitalization. In Polkadot, approximately 30,000 nominators maintain preference lists drawn from around 1,500 validator candidates, each containing up to 16 entries. The election itself is governed by a variant of the Phragm\'{e}n algorithm \citep{BrillEtAl2024, 10.1145/3479722.3480988}, which determines both which validators become active and how stakes are allocated from nominators to validators. As Polkadot is permissionless and offers pseudoanonymity, a single real entity might operate multiple accounts (called ``stashes") that hold stake. Each of these stashes may carry its own preference list and stake allocation. This practice of maintaining multiple accounts is fairly common because it enables nominators to diversify exposure across different sets of validators, mitigating the risk of losing funds due to slashes or other failed operations.

The problem that a nominator with multiple stashes faces is, at its core, a \emph{portfolio selection} problem. The nominator must decide which validators to include in each stash and how to distribute funds across them, knowing that the ultimate allocation of validators to stashes is determined by an election process whose outcome depends on the collective nominations of all participants and is therefore uncertain from any individual nominator's perspective. This creates a complex decision environment where profitability and diversification must be balanced under uncertainty.

Despite the economic significance and complexity of this decision, nominators currently receive little formal support. Existing decision aids in the blockchain domain have primarily focused on evaluating individual validators or blockchain platforms through multi-criteria methods \citep{GorccunEtAl2023,KrishankumarEtAl2024,BuyukozkanEtAl2021,WenLiao2024}, rather than on the portfolio-level problem of jointly optimizing the composition of multiple stashes. Meanwhile, the governance literature has identified several recurring pitfalls in blockchain ecosystems, including the lack of structured decision support for participants \citep{VanHaaren2023}. In a prior study, \citet{GehrleinEtAl2023} developed a decision support system based on active preference learning and multi-attribute value theory to help nominators rank individual validators according to their preferences. That system was validated with 115 real nominators from the Polkadot ecosystem and demonstrated improvements over unaided selection in terms of cognitive effort, preference accuracy, and staking performance. However, the approach was limited to evaluating and ranking individual validators and did not address the portfolio-level question of how to optimally distribute them across multiple stashes. This paper extends that line of research from the selection of individual validators to the multi-objective optimization of entire nomination portfolios.



In general, multi-objective portfolio optimization has traditionally been developed in the context of financial portfolios, where the classical return--risk trade-off has been generalized to account for additional objectives, constraints, and uncertainty. Recent reviews, however, emphasize that portfolio optimization has expanded far beyond its financial origins and now covers a wide range of selection and allocation problems in which \acp{DM} must form sets of alternatives under competing performance criteria \citep{SaloEtAl2024}. In the financial setting, this line of research includes deterministic and stochastic multiple-objective programming formulations of portfolio selection \citep{MasmoudiAbdelaziz2018}, as well as deterministic mean--variance models enriched with realistic constraints and alternative solution techniques \citep{KalayciEtAl2019}. From a computational perspective, population-based approaches have also played an important role, especially when complex constraints, non-convex feasible regions, or several conflicting objectives make the explicit construction of efficient portfolios difficult; this development is surveyed, among others, by \citet{MetaxiotisLiagkouras2012} in the context of multi-objective evolutionary algorithms for portfolio management. In the broader decision aiding literature, portfolio decision analysis (PDA) studies the selection of subsets of projects, actions, or resources while accounting for constraints, preferences, interactions, and uncertainties \citep{SaloKeislerMorton2011,LiesioEtAl2021}. This perspective is relevant for our study because the recommended object is not a single validator, but a structured set of nominations whose consequences depend on multiple criteria, the \ac{DM}'s preferences, and the allocation process.

A particularly relevant stream of this literature concerns portfolio problems in which the selection of high-level objects is coupled with the assignment or selection of lower-level elements. In competence-driven project portfolio selection, \citet{GutjahrEtal2008} combined project selection, scheduling, and staff assignment, while explicitly accounting for the development of human competencies over time. This line was further extended to multi-objective formulations, in which economic performance is considered alongside competence-oriented objectives \citep{GutjahrEtal2010}, and to uncertain settings, where staff assignment and project portfolio selection are treated within a bi-objective framework \citep{GutjahrReiter2010}. More recently, \citet{NoroDias2023} proposed a bi-objective project portfolio model in which projects are selected jointly with the allocation of agents, seeking to balance economic gains with the development of agents' skills. These works are closely aligned with our setting because they recognize that portfolio quality may depend not only on which main objects are selected, but also on how associated resources, agents, or components are assigned. The recent portfolio-of-portfolios model of \citet{BarbatiGrecoFigueira2025} makes this relationship explicit by considering the joint selection of a portfolio of projects and portfolios of elements assigned to them, with qualitative and quantitative requirements imposed on the elements. Our setting shares this multi-level view of portfolio construction, but differs in that the final allocation is shaped by the specific election and assignment mechanisms of proof-of-stake blockchains. 

In this context, our contribution is both methodological and application-oriented. First, we introduce a new portfolio collection optimization problem motivated by the practical setting of proof-of-stake blockchains. Instead of recommending validators independently, we model the decision faced by a nominator who operates several stashes and must construct a collection of nomination portfolios, one for each stash. 
We represent a nomination as a weighted assignment of validators to stashes and model the allocation outcome probabilistically. This leads to a portfolio problem that differs from standard financial portfolio selection and from classical project portfolio models, because the final exposure is induced by a stochastic allocation mechanism rather than chosen directly.

Second, we propose a bi-objective formulation that makes the trade-off underlying the nominator's decision explicit. The first objective maximizes the expected utility of the allocated validators and thus captures portfolio quality and profitability from the perspective of the nominator. The other objective maximizes the expected entropy of the allocation and thus captures diversification across validators and stashes. In this way, risk mitigation is not introduced through an exogenous penalty or a simple cardinality constraint, but through an allocation-sensitive measure that reflects how the nominator's stake is expected to be distributed after the blockchain election. The resulting model provides a transparent way to distinguish portfolios that concentrate exposure on highly valued validators from portfolios that sacrifice part of this expected quality in exchange for broader diversification.

Third, we develop a comprehensive decision support framework to solve this problem in a personalized way. Validator utilities are obtained via an active preference learning procedure based on multi-attribute value theory, extending previous work on validator selection by adapting the learning criterion to the portfolio-construction setting~\citep{GehrleinEtAl2023}. These learned utilities are then integrated into the bi-objective optimization model, which is solved with a multi-objective evolutionary algorithm. Since evaluating candidate portfolios is computationally demanding, we also study several solution strategies that combine exact evaluation, restricted candidate sets, and approximate evaluation, aiming to obtain high-quality Pareto fronts with short interaction time.

Fourth, we address the problem of selecting a final portfolio from the Pareto front. Although a two-objective front is easier to visualize than a higher-dimensional trade-off, the meaning of the two axes -- expected utility and expected entropy -- may still be abstract for nominators. We therefore introduce an interactive navigation procedure that presents portfolios in the decision space and asks the nominator to move along the Pareto front toward higher expected utility or higher diversification. The procedure follows a binary-search logic and reduces the number of comparisons needed to identify a satisfactory portfolio, while preserving the nominator's control over the final trade-off.

Finally, we evaluate the approach in two complementary ways. Numerical simulations are used to assess the revised preference learning metric and the computational strategies for bi-objective optimization. The empirical part of the study involves experienced Polkadot nominators who actively stake with multiple stashes. Their assessment provides evidence on the cognitive demands of the interaction, the perceived quality of the recommended portfolios, and the practical usefulness of the proposed decision support system. This validation is important because the problem is not only computationally complex, but also embedded in a real decision environment in which users have established practices, individual risk attitudes, and domain-specific knowledge.



The remainder of the paper is organized as follows. Section~\ref{sec:problem} introduces the formal setting, including the assignment of validators to stashes and the allocation process. Section~\ref{sec:portfolio} formulates the bi-objective optimization problem, develops the two objective functions, and describes the interactive portfolio selection procedure. Section~\ref{sec:experiments} presents the experimental results, including simulations and the empirical evaluation with real nominators. The last section concludes and discusses avenues for future research.













\section{Preliminaries and problem statement}
\label{sec:problem}
We consider the problem faced by a nominator with possibly multiple identities/accounts, each of which can be used to nominate validators.
Using multiple accounts is common because it enables nominators to reduce risk through diversification. Furthermore, the case of a nominator having a unique account is a special instance of the model that we shall develop and, therefore, will fall under the more general umbrella proposed by our formalization.
In this context, we assume the existence of a finite non-empty set of validators $V=\{v_1,\ldots,v_m \}$. Then, we consider that, when it comes to compiling shortlists of validators, a nominator with possibly multiple accounts $(n\ge 1)$ faces a problem in which they have $n$ stashes $B_1,\ldots,B_n$, each of which may contain at most $C$ validators. The same validator may appear in more than one stash. In the special case of $n=1$, the problem collapses into the choice of a subset of $V$ with cardinality at most $C$.

A possible \emph{assignment} of this type can be represented by a matrix $\mathbf{X}=(x_{ij})_{m \times n}$ with $x_{ij} \in \{ 0,1 \}$ and $\sum_{i=1}^{m} x_{ij} \leq C$ for all $j$, where 
\[
x_{ij}=
\begin{cases}
1,& \text{if validator $v_{i}$ is assigned to (``put into'') $B_j$}, \\
0, & \text{otherwise.}
\end{cases}
\]
We call $\mathcal{X}_{m,n,C}$ the set of all these matrices, i.e.,
\[
\mathcal{X}_{m,n,C} = \left\{ \mathbf{X}\, \bigg| \, \mathbf{X} \in \{0,1 \}^{m \times n}, \, 1 \le \sum_{i=1}^{m}x_{ij} \leq C,\; \forall j \right\}.
\]

\begin{example}
\label{ex:1}
Given a set of validators $V=\{ v_1,\ldots,v_6\}$, with three identities corresponding to the three stashes $B_1, B_2, B_3$ of maximum capacity $4$, i.e. $C=4$, then the following matrix
\begin{equation}
\label{eq:assignment}
\mathbf{X} = (x_{ij})_{6 \times 3} = \bordermatrix{~ & B_1 & B_2 & B_3 \cr
              v_1 & 0 & 0 & 1\cr
              v_2 & 0 & 0 & 1\cr
              v_3 & 0 & 0 & 1\cr
              v_4 & 1 & 1 & 0\cr
              v_5 & 0 & 0 & 0\cr
              v_6 & 0 & 1 & 0\cr}
              \in \mathcal{X}_{6,3,4}
\end{equation}
is an example indicating that $B_1 = \{ v_4 \} $, $B_2 = \{ v_4 , v_6\}$ , $B_3 = \{ v_1, v_2 , v_3 \}$.
\end{example}
Note that the total number of possible assignments of validators to stashes corresponds to the cardinality of the set $\mathcal{X}_{m,n,C}$ and is equal to
\begin{equation}
|\mathcal{X}_{m,n,C}|=
\left( \sum_{k=1}^{C}  \binom{m}{k} \right)^{n}.
\end{equation}
In non-trivial applications, such a~number grows very quickly. In particular, for Example~\ref{ex:1} it is 175,616.

The nomination process, whose outcome is a matrix $\mathbf{X}$, is followed by an \emph{allocation} process, which will be discussed later in greater detail. In brief, the allocation process selects at most one nominated validator from each stash. Then, the nominator's gain is linked to the performance of the validators allocated to their stashes. We use the notation $(v_i \mapsfrom B_j)$ to state that the validator $v_i$ was allocated from stash $B_j$. If we consider Example \ref{ex:1}, then
\begin{equation}
\label{eq:allocation}
    (v_4 \mapsfrom B_1) \hspace{0.2cm} (v_4 \mapsfrom B_2) \hspace{0.2cm} (v_2 \mapsfrom B_3)
\end{equation} is a feasible allocation, compatible with the illustrative assignment \eqref{eq:assignment}.

One further source of complexity is that the blockchain environment can accommodate only a subset of the candidate validators in each period. The validators chosen to run the blockchain during a given period are called \emph{active}. The process that determines what validators are active in the next period and how they are allocated from the stashes is called \emph{election} and is deterministic. For example, in our case study, it is based on the Phragm\'{e}n algorithm~\citep{BrillEtAl2024}, whose output depends on the composition of all the stashes of all the nominators. Although the output of the algorithm is deterministic (unless ties are present), given its complexity and the lack of information that any nominator has on the nominations of other nominators, we shall model its outcome as a random variable.

\begin{figure}[h]
    \centering
    \includegraphics[width=0.6\textwidth]{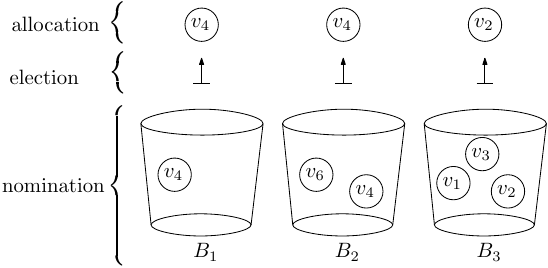}
    \caption{Representation of the nomination phase \eqref{eq:assignment} and subsequent allocation phase \eqref{eq:allocation}.}
    \label{fig:bins}
\end{figure}

Figure \ref{fig:bins} shows a graphical interpretation of the nomination and allocation processes. In the nomination stage, the nominator assigns validators to the available stashes, and the same validator may appear in more than one stash. In the subsequent allocation stage, the election mechanism selects at most one validator from each stash among those that are active. For the capital to earn rewards, at least one of the allocated validators must become active in the subsequent round; otherwise, the capital sits idle. Only then can a nominator realize profits (or losses) on that stash. The figure also illustrates that the final exposure of the nominator may differ from the initial nomination: several nominated validators may not be allocated, while the same validator can receive stake from more than one stash if it is allocated in multiple stashes.

After the allocation phase, rewards (in the form of the blockchain's native tokens) of nominators are dependent on the actions/behavior of the validators allocated to their stashes. This serves as an incentive for nominators to nominate validators they deem sufficiently trustworthy, and thus, through the election process, to guarantee a high level of reliability across the entire blockchain.

So far, we assumed that all stashes have the same weight, meaning that the nominator's capital has been split equally among them. More generally, however, different stashes may contain different amounts of stake. Therefore, we couple $\mathbf{X}$ with a weight vector $\mathbf{w}=(w_{1},\ldots,w_{n})$, which represents the proportions of investment assigned to each stash. For example, if the allocation illustrated in Figure~\ref{fig:bins} is combined with weights $\mathbf{w}=(w_1,w_2,w_3)$, then validator $v_4$ receives $w_1+w_2$ units of stake and validator $v_2$ receives $w_3$. In particular, we have $\mathbf{w} \in \mathcal{W}_{n}$ where
\begin{equation}
\mathcal{W}_{n} = \{  \mathbf{w} \;|\; w_{i}\geq 0, \, w_{1}+\cdots+ w_{n} = 1 \}
\end{equation}
represents the $(n-1)$-th dimensional unit simplex. Hereafter, a weighted nomination of validators can be unambiguously represented by a pair $(\mathbf{X},\mathbf{w})$ and our problem concerns its optimal selection.

\section{A portfolio optimization problem in proof-of-stake blockchains}
\label{sec:portfolio}
We are interested in developing a number of quantitative metrics that can be used as objective functions $f_{1},\ldots,f_{q}: \mathcal{X}_{m,n,C} \times \mathcal{W}_{n} \rightarrow \mathbb{R}$, to assess the goodness of the nomination presented by a given nominator. Then, our final goal is to formulate and solve an optimization problem to find the 
Pareto efficient set of nominations. Hereafter, for simplicity, we will often drop the subscripts of $\mathcal{X}_{m,n,C}$ and $ \mathcal{W}_{n}$.
We first examine the allocation process in more detail, which --- based on the nomination of each nominator --- matches the shortlists presented by each nominator to some validators. When validator $v_i$ is selected from a stash $B_j$, we say that $v_i$ is \emph{allocated} to $B_j$. 
If we want to formulate an optimization problem, we must mimic the allocation process with sufficient reliability.

\subsection{Allocation probabilities}

Each validator $v_i \in V$ has an estimated probability $p_{\mathrm{act}}(v_i)\in[0,1]$ of being active in the next period. This estimation can be obtained from historical activity, time-series analysis, or subjective forecasting. Since the nominator does not observe the full set of nominations submitted by all other participants, the complete joint distribution of the future active validator set is not available. We therefore model the activation status of each nominated validator as an independent Bernoulli random variable, conditional on the information available to the nominator. This simplifying assumption allows approximating the probability of each active subset while retaining the marginal activation probabilities of individual validators.

For a stash $B_j$, the probability that exactly the subset of validators $A\subseteq B_j$ is active in the next period is approximated as
\begin{equation}
\Pr(A\mid B_j)
=
\prod_{v_h\in A} p_{\mathrm{act}}(v_h)
\prod_{v_k\in B_j\setminus A} \left(1-p_{\mathrm{act}}(v_k)\right),
\end{equation}
where $v_i\in B_j$ means that validator $v_i$ is nominated in the $j$-th stash ($x_{ij}=1$). This approximation should not be interpreted as a claim that the blockchain election mechanism generates independent activation events; rather, it is a tractable representation of the nominator's uncertainty based on individual activation probabilities.

\begin{example}
\label{ex:2}
Consider Example \ref{ex:1}, and stash $B_3$, in which the validators $v_1, v_2$, and $v_3$ were nominated. If we further assume that the probabilities that they are going to be active are $p_{\mathrm{act}}(v_1)=0.5$, $p_{\mathrm{act}}(v_2)=0.4$, and $p_{\mathrm{act}}(v_3)=0.3$, respectively, then the probabilities that each subset of $B_3$ is active are
\begin{align*}
 \Pr(\emptyset | B_3) &= (1-0.5)(1-0.4)(1-0.3)=0.21, \\
 \Pr( \{ v_1 \} | B_3) &= (0.5)(1-0.4)(1-0.3)=0.21,\\
 \Pr( \{ v_2 \} | B_3) &= (1-0.5)(0.4)(1-0.3)=0.14,\\
 \Pr( \{ v_3 \} | B_3) &= (1-0.5)(1-0.4)(0.3)=0.09,\\
 \Pr( \{ v_1,v_2 \} | B_3) &= (0.5)(0.4)(1-0.3)=0.14, \\
 \Pr( \{ v_1,v_3 \} | B_3) &= (0.5)(1-0.4)(0.3)= 0.09,\\
 \Pr( \{ v_2,v_3 \} | B_3) &= (1-0.5)(0.4)(0.3)= 0.06,\\
 \Pr( \{ v_1,v_2,v_3 \} | B_3) &= (0.5)(0.4)(0.3)= 0.06. 
\end{align*}
\end{example}
For \(v_i\) to be allocated from \(B_j\), two events must occur: (i) a subset $A \subseteq B_j$ of validators must be active with the condition that $v_i \in A$, and (ii) out of all validators in $A$, the stakes are allocated to $v_i$. Formally, we can write
\begin{equation}
\label{eq:probability}
\Pr(v_{i} \mapsfrom B_j) = \sum_{A \subseteq B_j | v_{i }\in A}  {\Pr(v_{i} | A)}{\Pr(A|B_j)}.
\end{equation}
If we turn our attention to the probabilities ${\Pr(v_{i} | A)}$ for individual validators, in the absence of further information, we could follow a naive approach and assume the probability that a given validator is allocated from $B_j$ is uniformly distributed among all the validators in $A$, i.e. 
\begin{equation}
\label{eq:uniform}
    \Pr (v_i | A )=\frac{1}{|A|}.
\end{equation}
\begin{example}
    Based on the probabilities and subset cardinalities from Example \ref{ex:2}, we have 
    \begin{align*}
 \Pr(v_{1} \mapsfrom B_3) & = 0.21 + \frac{1}{2}\cdot0.14 + \frac{1}{2}\cdot0.09 + \frac{1}{3}\cdot0.06 = 0.345,\\
  \Pr(v_{2} \mapsfrom B_3) & = 0.14 + \frac{1}{2}\cdot0.14 + \frac{1}{2}\cdot0.06 + \frac{1}{3}\cdot0.06 = 0.26,  \\
  \Pr(v_{3} \mapsfrom B_3) & = 0.09 + \frac{1}{2}\cdot0.09 + \frac{1}{2}\cdot0.06 + \frac{1}{3}\cdot0.06 = 0.185. &  
\end{align*}
The empty-set case completes the probability mass. It represents the scenario in which no validator in $A$ is active, and therefore no validator is allocated from $B_j$. In the example, it corresponds to $\Pr(\emptyset \mapsfrom B_j)= 0.21$.
\end{example}

A more informed estimator can exploit information about each validator's total nomination stake. Let $t(v_i)$ denote the total stake nominated to validator $v_i$, i.e., the sum of the stakes of all nominators supporting this validator. Under the Polkadot election mechanism, validators with larger total nomination stakes tend to be less likely to receive an additional allocation from a specific nominator, because their support is already spread across many nominators. Conversely, validators with smaller total nomination stakes are relatively more likely to be assigned when they belong to the active set. We therefore approximate the conditional probability that validator $v_i$ is selected from the active subset $A$ as inversely proportional to its total nomination stake:\begin{equation}
\label{eq:non_naive}
\Pr(v_i | A) = \frac{1/t(v_i)}{\sum_{v_k \in A} 1/t(v_k)}.
\end{equation}

\begin{example}
    Assume that an active set $A = \{v_1,v_2,v_3 \}$ consists of three validators with the same total stake $t(v_1)=t(v_2)=t(v_3) = 200$. Then, the probability of being allocated is the same:
    \[
    \Pr(v_i | A) = \frac{1/t(v_i)}{\sum_{v \in A} 1/t(v)} = \frac{0.005}{0.015} = \frac{1}{3} \approx 0.333.
    \]
    If, instead, the total stakes were $t(v_1)=200$, $t(v_2)=300$, and $t(v_3)=400$, then the estimated probabilities that $v_1$, $v_2$, and $v_3$ are allocated from $A$ are
    \begin{equation*}
 \Pr(v_{1}|A)  = \frac{0.005}{0.0108} = 0.462, \hspace{0.5cm} 
  \Pr(v_{2}|A)  = \frac{0.00333}{0.0108} = 0.308, \hspace{0.5cm} 
   \Pr(v_{3}|A)  = \frac{0.0025}{0.0108} = 0.231.
\end{equation*}
\end{example}

{
To assess the quality of the estimation via \eqref{eq:non_naive}, a sample of over 3,500,000 historical allocations $\Pr(v_{i} \mapsfrom A)$ was extracted out of which $80\%$ were used as training data, $10\%$ as validation, and the remaining $10\%$ as test data. Each selection was mapped to a vector $s$ with individual elements
\[
s_{j}=
\begin{cases}
1,& \text{if $i = j$}, \\
0, & \text{otherwise},
\end{cases}
\]
and each estimation to a vector $e$, where $e_j$ denotes the estimated probability that $v_j$ is selected. Then, \ac{MAE} was used to evaluate the performance of the different algorithms. The neural network model achieved the best result, with an \ac{MAE} of $0.0861$. The proposed approach followed closely, obtaining an \ac{MAE} of $0.0866$. In contrast, both naive estimators performed noticeably worse. The random approach reached an \ac{MAE} of $0.0884$, while the equal-probability solution achieved an \ac{MAE} of $0.0883$.


The results indicate that the stake-based estimator in \eqref{eq:non_naive} captures a substantial part of the allocation mechanism. Its MAE is very close to that of the neural network model, with an absolute difference of only $0.0005$, and it improves over both the equal-probability and random baselines. This suggests that the inverse relationship between a~validator's total nomination stake and its probability of being allocated provides a meaningful approximation of the underlying process. At the same time, the small advantage of neural networks comes at the cost of reduced interpretability, greater implementation complexity, and greater dependence on available historical training data. Since our goal is not only predictive accuracy but also a transparent decision support model that can be explained to nominators, we favor the stake-based estimator. Hence, in the remainder of the paper, we use \eqref{eq:probability} together with \eqref{eq:non_naive} to estimate $\Pr(v_i \mapsfrom B_j)$.

\subsection{A utility-based measure of portfolio performance} Similarly to what happens in a standard portfolio optimization problem, a nominator is interested in maximizing the reward coming from its investment. To this end, a utility value can be associated with each validator. That is, if we assume that these values are normalized, a nominator could define a function $U: V\rightarrow [0,1]$ such that $U(v_{i})$ is the utility of the $i$-th validator. 
Manually defining the parameter values of $U$ could require an excessive effort due to (i) the large number of validators and (ii) the multiplicity of attributes of validators that can be used as indicators of their reliability. Multi-attribute decision analysis provides a natural way to address this difficulty. This study builds on the methodology based on multi-attribute value theory, which was already validated in the Polkadot blockchain environment~\citep{GehrleinEtAl2023}. 

In particular, an active preference learning approach was used to estimate the marginal value functions $u$ of nominators with respect to five relevant criteria $g_j$. Attribute levels of validators were then evaluated according to these marginal value functions and aggregated additively so that each validator $v_i$ was mapped into a utility value $U(v_i) = \sum u_j(g_j(v_i)) \in [0,1]$, where $g_j(v)$ denotes the performance of validator $v$ on criterion $j$. Higher values of the utility indicate better validators. As the UTA \citep{JacquetLagreze1982} method is applied, the functions $u_j$ take the form of piecewise linear functions with predetermined characteristic points at which the function breaks. Examples of such functions are presented in Figure~\ref{fig:utility}.

\begin{figure}[h!]
    \centering
\includegraphics[width=\linewidth]{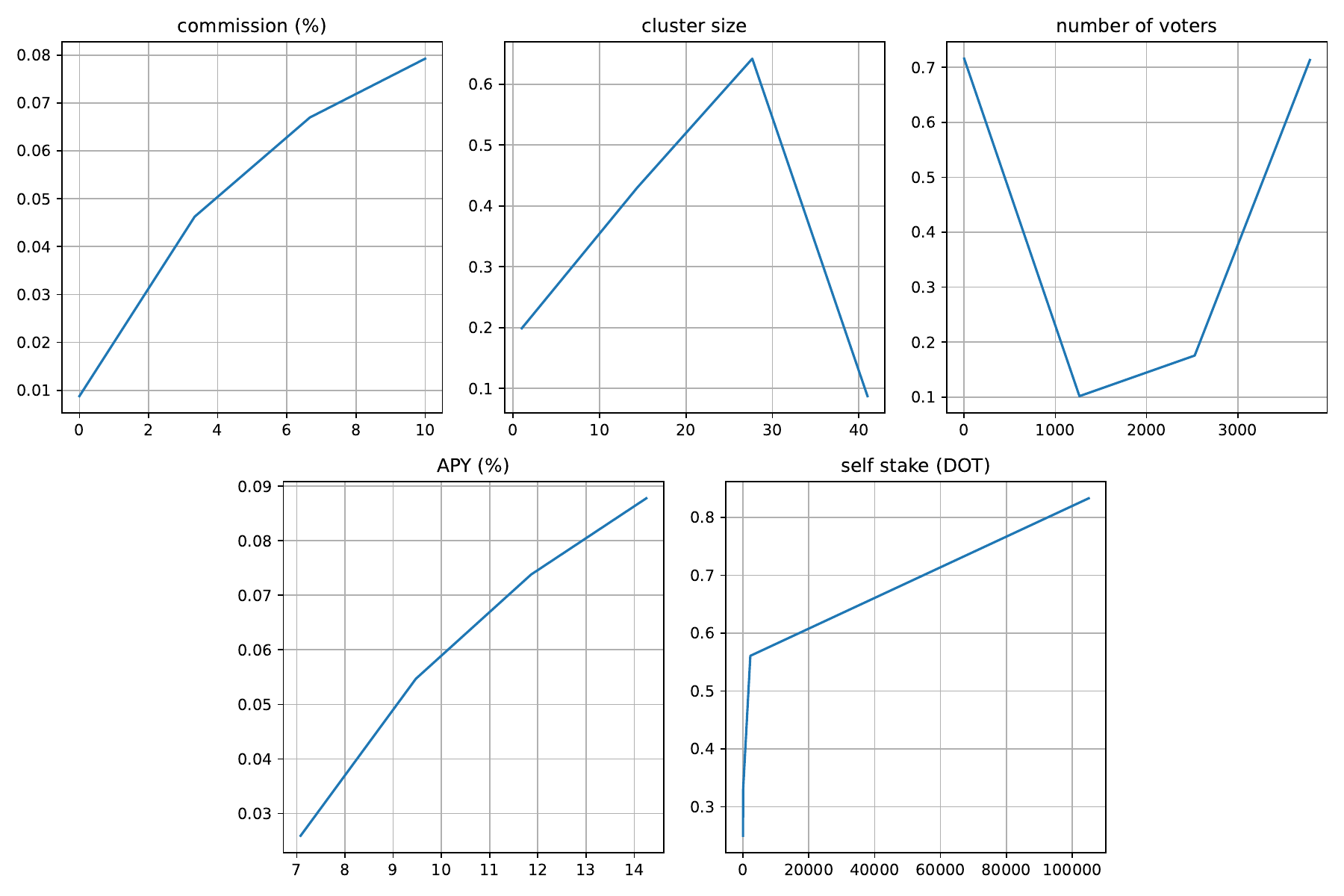}
    \caption{Example utility functions obtained using the UTA method with active learning. The criteria cluster size and number of voters are non-monotonic, while the remaining three are gain-type criteria.}
    \label{fig:utility}
\end{figure}

The main idea is to iteratively identify a pairwise elicitation question that maximizes information gain, present it to the \ac{DM}, and update the model based on the response. This procedure can be repeated a selected number of times or until the desired model quality is achieved. The information gain is calculated by simulating both possible answers to a given question and evaluating the model's quality after each potential update. To this end, the space of possible utility functions is sampled multiple times; for each sample, a ranking of alternatives is generated. The model quality is then assessed based on the minimal Spearman rank correlation coefficient between each pair of these rankings. After each question is answered by the \ac{DM}, the space of possible utility functions is restricted to those consistent with the provided preferences. Specifically, if the \ac{DM} indicates that an alternative $a$ is preferred over an alternative $b$, then only utility functions scoring higher $a$ than $b$ are retained, i.e., $U(a) > U(b) \iff \sum u_j(g_j(a)) > \sum u_j(g_j(b))$.

The main difference compared to the previous study~\citep{GehrleinEtAl2023} is the increased focus on the top positions in the rankings, as the goal of the optimization process is to construct the best possible portfolio based on the highest-ranked alternatives. To account for this change, a different metric for evaluating the consistency of a set of models was proposed. The previously used Spearman rank correlation coefficient was replaced with a Pearson correlation coefficient computed on logarithmically transformed ranks. This choice follows the common idea in preference learning and information retrieval that discrepancies at the top of a ranking should be penalized more strongly than discrepancies near the bottom~\citep{FuernkranzHuellermeier2016PreferenceLearningRanking}. In particular, weighted ranking errors and discounted cumulative gain measures use logarithmic discounting to give higher importance to top-ranked elements~\citep{JarvelinKekalainen2002}. Analogously, applying a logarithmic transformation to ranks compresses differences among lower-ranked validators while preserving greater sensitivity to changes among the highest-ranked ones, which are the most relevant for portfolio construction.

The additional complexity of the setting considered in this paper stems from the non-deterministic nature of the allocation phase's output. We assume that we are interested in maximizing the \emph{expected utility} of the assignment $(\mathbf{X},\mathbf{w})$. The expected utility of a stash $B_j$, which we call $\mathbb{E}[\Pi(B_j)]$, is the weighted sum of the utilities of the validators in the stash, and the weights are the probabilities of being chosen from that stash, i.e.,
\begin{equation}
\mathbb{E}[\Pi(B_j)] =  \sum_{v_i \in B_j}  U(v_{i}) \cdot \Pr(v_{i} \mapsfrom B_j).
\end{equation}
Then, as the \emph{first objective function}, we consider a weighted assignment of validators to stashes, represented by $ ( \mathbf{X}, \mathbf{w} )$, we have
\begin{equation}
\label{eq:f1}
\tag{{OBJ-1}}
\mathbb{E}[\Pi(\mathbf{X},\mathbf{w})] = \sum_{j=1}^{n} w_{j} \sum_{v_i \in B_j}  U(v_{i}) \cdot \Pr(v_{i} \mapsfrom B_j),
\end{equation}
where \(w_j\) is the proportion of stake assigned to the \(j\)th stash. In summary, the weights $\mathbf{w}$ determine the contributions of the validators allocated from different stashes and the assignment of validators to stashes described by $\mathbf{X}$ acts on the probabilistic term $\Pr(v_i \mapsfrom B_j)$ and makes (OBJ-1) nonlinear.

\begin{example}
Consider four validators $V=\{v_1,v_2,v_3,v_4\}$ with the following utilities
\[
U(v_1)=0.90,\qquad U(v_2)=0.80,\qquad U(v_3)=0.65,\qquad U(v_4)=0.50.
\]
Suppose that a nominator uses two stashes,
\[
B_1=\{v_1,v_2\},\qquad B_2=\{v_2,v_3,v_4\},
\]
and splits the stake according to $\mathbf{w}=(0.7,0.3)$. Assume that, using the allocation model described above, the following allocation probabilities are obtained:
\[
\begin{array}{c|ccccc}
 & v_1 & v_2 & v_3 & v_4 & \emptyset\\
\hline
B_1 & 0.60 & 0.30 & - & - & 0.10\\
B_2 & - & 0.20 & 0.50 & 0.20 & 0.10
\end{array}
\]
where $\emptyset$ denotes the event that no validator is allocated from the corresponding stash. This event contributes zero to the expected utility.
The expected utility of the first stash is therefore
\[
E[\Pi(B_1)]
=
0.90\cdot 0.60 + 0.80\cdot 0.30
=
0.78,
\]
whereas for the second stash, we obtain
\[
E[\Pi(B_2)]
=
0.80\cdot 0.20 + 0.65\cdot 0.50 + 0.50\cdot 0.20
=
0.585.
\]
Taking into account the distribution of stake across the two stashes, the expected utility of the weighted nomination is
\[
E[\Pi(\mathbf{X},\mathbf{w})]
=
0.7\cdot 0.78 + 0.3\cdot 0.585
=
0.7215.
\]
This example illustrates that the utility of a nomination depends jointly on the utilities of validators, their probabilities of being allocated from each stash, and the weights assigned to the stashes. In particular, the contribution of the same validator may differ across stashes because its allocation probability is stash-dependent, and validators in stashes with larger stake weights have a stronger effect on the overall expected utility.
\end{example}




\subsection{An entropy-based measure of diversification} 
A nominator may wish to diversify the investment across validators rather than concentrate the entire exposure in a small subset. Consider the stashes $B_1,\ldots,B_n$. From a diversification perspective, it is desirable that the allocation phase results in different validators being assigned to different stashes, preferably with comparable amounts of stake. At the opposite extreme, diversification is minimal when all allocations from all stashes point to the same validator, effectively concentrating the nominator's stake on a single actor. The degree of diversification depends on both the structure of the nomination and the uncertain allocation outcome. If the stashes are disjoint, then any feasible allocation from these stashes necessarily involves distinct validators, provided that one validator is allocated from each stash. Conversely, if all stashes contain one and the same validator, then every successful allocation leads to the same validator, and no diversification is achieved. Between these two extreme cases, the final allocation cannot be determined with certainty before the election. It is therefore natural to assess diversification in probabilistic terms, by considering the distribution of stake induced by all possible allocation outcomes.

If we consider $n$ stashes and a possible allocation of validators from those stashes, we note that such allocation can be represented by a list $\mathbf{e}= ( {e}_1,\ldots,{e}_{n} ) \in (V\cup \emptyset )^{n}$ together with the vector $\mathbf{w}$. Clearly, some elements of the list can be the same. From the lists $\mathbf{e}$ and $\mathbf{w}$, we can associate each validator $v_i$ with a value $\omega(v_i) \geq 0$ which represents the total amount of stashes allocated to $v_i$. Now, considering only validators with positive $\omega(v_i)$, we propose the entropy of the stash distribution as
\begin{equation}
H(\mathbf{e},\mathbf{w}) = - \sum_{\substack{v_i \in V \\ \omega(v_i)>0}} \omega(v_i) \log_2 \omega(v_i).
\end{equation}

Later, in our implementation, we consider only assignments of validators to stashes for which the probability that no validator is assigned to any stash is near zero. By doing so, we ensure that, for each feasible allocation $\mathbf{e}$, we have $\sum_{v_i \in V} \omega(v_i) \approx 1$, and that no significant distortion applies to the entropy-like formulation of $H$.

\begin{example}
Consider the allocation of validators $\mathbf{e}=(v_3,v_7,v_3,v_8)$ from four stashes with $\mathbf{w}=(0.4,0.3,0.2,0.1)$. Then, we have $\omega(v_3) = 0.4+0.2 = 0.6$, $\omega(v_7)=0.3$, $\omega(v_8)=0.1$. The entropy of this allocation is 
\[
H(\mathbf{e},\mathbf{w}) \approx 1.2955.
\]
Instead, if we consider the allocation of validators $\mathbf{e}=(v_2,v_2,v_2,v_2)$ from four stashes with $\mathbf{w}=(0.4 , 0.3, 0.2, 0.1)$, then we have $\omega(v_2)=1$ and minimum entropy, $H(\mathbf{e},\mathbf{w})=0$. Conversely, the allocation of validators $\mathbf{e}=(v_4,v_9,v_8,v_2)$ from four stashes with $\mathbf{w}=(1/4, 1/4, 1/4, 1/4)$ yields $\omega(v_4)=\omega(v_9)=\omega(v_8)=\omega(v_2)=1/4$ and maximum entropy $H(\mathbf{e},\mathbf{w})=2$.
\end{example}

At this point, we can consider the expected entropy of the nomination as a function of the pair $(\mathbf{X},\mathbf{w})$. That is, the \emph{second objective function} is
\begin{equation}
\label{eq:f2}
\tag{{OBJ}-2}
\mathbb{E}(H(\mathbf{X},\mathbf{w}))= \sum_{\mathbf{e} \in \mathcal{E}}\Pr(\mathbf{e})H(\mathbf{e},\mathbf{w}),
\end{equation}
where $\mathcal{E}$ is the set of possible allocations given the assignment $\mathbf{X}$. Clearly, each extraction corresponds to an allocation of stashes to validators. 
To compute the expected entropy, we approximate the probability of a complete allocation vector $\mathbf{e}=(e_1,\ldots,e_n)$ by assuming that allocations from different stashes are conditionally independent given the estimated stash-level allocation probabilities. Thus,
\begin{equation}
\Pr(\mathbf{e})
=
\prod_{j=1}^n \Pr(e_j \mapsfrom B_j),
\end{equation}
where $e_j\in V$ denotes the validator allocated from stash $B_j$.


\subsection{A bi-objective optimization problem}
\label{sub:bi-objective}


We assume that a nominator with multiple stashes is interested in the simultaneous maximization of the expected utility and the expected entropy of the allocation resulting from the pair $(\mathbf{X},\mathbf{w})$. Hence, considering the two objective functions \eqref{eq:f1} and \eqref{eq:f2} defined in the previous subsections, we state the optimization problem:
\begin{equation}
\label{eq:bi-objective}
\maximize_{\mathbf{X} \in \mathcal{X},\mathbf{w} \in \mathcal{W}} \quad \left\langle \, \mathbb{E}[\Pi(\mathbf{X},\mathbf{w})],\mathbb{E}[H(\mathbf{X},\mathbf{w})] \,  \right\rangle,
\end{equation}
whose objectives are nonlinear.

Having only two objectives, we adopt an \textit{a posteriori} approach and favor an explicit analysis of the trade-off between them. By doing so, we need to identify an approximation of all the non-dominated (economically relevant) solutions to the problem and present them to the nominator for scrutiny and final selection. To generate a finite set of such solutions, we employed a multi-objective evolutionary algorithm, NSGA-II \citep{DebEtAl2002, Pasandideh2015}, implemented in Pymoo \citep{BlankDeb2020} in Python. {Figure \ref{fig:Pareto} shows an example of a Pareto front for a nominator with three stashes associated with weights $\mathbf{w}=(0.75,0.15,0.10)$. 
}

\begin{figure}[h!]
    \centering
\includegraphics[width=0.6\linewidth]{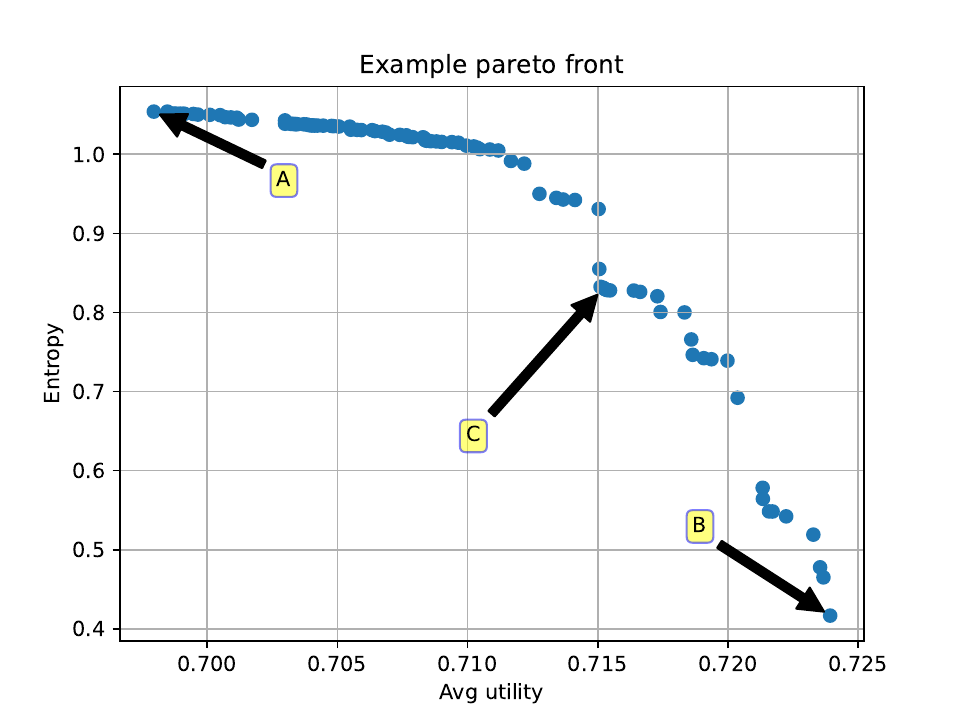}
    \caption{An instance of the Pareto front for the bi-objective optimization problem \eqref{eq:bi-objective}. Each point represents a portfolio characterized by a pair $(\mathbf{X},\mathbf{w})$.}
    \label{fig:Pareto}
\end{figure}

For interpretability, we highlight three representative nondominated solutions, denoted by A, B, and C, in Figure~\ref{fig:Pareto}. Table~\ref{tab:portfolios} reports these solutions in the decision space, i.e., as collections of nominations assigned to individual stashes. They illustrate three qualitatively different ways of balancing expected utility and diversification:

\begin{description}
\item[Point A] corresponds to a highly diversified portfolio collection. In Table \ref{tab:res1}, each validator appears in only one stash, so the final allocation is spread across distinct validators whenever one validator is allocated from each stash. Since the first stash has the largest weight, it contains validators with the highest utilities, while the remaining stashes are filled with different validators to preserve diversification. This solution, therefore, favors the entropy objective while still assigning the most valuable validators to the most important stash.

\item[Point B] corresponds to a performance-oriented portfolio collection. In Table \ref{tab:res3}, each stash contains the same set of top-utility validators, subject to the assumed minimum stash size. This increases the probability that a high-utility validator is allocated from each stash, but it also concentrates the nominator's exposure: the same validator may be allocated from several stashes. Thus, this solution prioritizes expected utility, while diversification plays only a minor role.

\item[Point C] represents an intermediate trade-off between the two objectives. In Table \ref{tab:res2}, the best validators appear in multiple stashes, which helps maintain a high expected utility. However, they are combined with additional validators to avoid the complete concentration observed in Point B. This solution therefore illustrates how the model can generate compromise portfolios that retain strong expected performance while improving the distribution of stake across validators.
\end{description}
\begin{table}[ht]
\centering
\caption{Comparison of the three representative solutions in Figure \ref{fig:Pareto}. Here, each point is represented in the decision space as a portfolio of stashes.}
\label{tab:portfolios}

\begin{subtable}{\textwidth}
\centering
\caption{Solution A}
\label{tab:res1}
\resizebox{0.8\columnwidth}{!}{
\begin{tabular}{c c c @{\hspace{0.8cm}} c c c @{\hspace{0.8cm}} c c c}
\toprule
\multicolumn{3}{c}{Stash 1} &
\multicolumn{3}{c}{Stash 2} &
\multicolumn{3}{c}{Stash 3} \\
\cmidrule(r){1-3} \cmidrule(r){4-6} \cmidrule(r){7-9}
Val ID & probability & utility &
Val ID & probability & utility &
Val ID & probability & utility \\
\midrule
1 & 0.15 & 0.78 & 4 & 0.22 & 0.7 & 6 & 0.72 & 0.61 \\
2 & 0.12 & 0.76 & 5 & 0.63 & 0.63 & 10 & 0.09 & 0.56 \\
3 & 0.73 & 0.71 & 7 & 0.15 & 0.6 & 11 & 0.19 & 0.55 \\
\bottomrule
\end{tabular}
}
\end{subtable}

\vspace{0.6cm}

\begin{subtable}{\textwidth}
\centering
\caption{Solution B}
\label{tab:res3}
\resizebox{0.8\columnwidth}{!}{
\begin{tabular}{c c c @{\hspace{0.8cm}} c c c @{\hspace{0.8cm}} c c c}
\toprule
\multicolumn{3}{c}{Stash 1} &
\multicolumn{3}{c}{Stash 2} &
\multicolumn{3}{c}{Stash 3} \\
\cmidrule(r){1-3} \cmidrule(r){4-6} \cmidrule(r){7-9}
Val ID & probability & utility &
Val ID & probability & utility &
Val ID & probability & utility \\
\midrule
1 & 0.15 & 0.78 & 1 & 0.15 & 0.78 & 1 & 0.15 & 0.78 \\
2 & 0.12 & 0.76 & 2 & 0.12 & 0.76 & 2 & 0.12 & 0.76 \\
3 & 0.73 & 0.71 & 3 & 0.73 & 0.71 & 3 & 0.73 & 0.71 \\
\bottomrule
\end{tabular}
}
\end{subtable}

\vspace{0.6cm}

\begin{subtable}{\textwidth}
\centering
\caption{Solution C}
\label{tab:res2}
\resizebox{0.8\columnwidth}{!}{
\begin{tabular}{c c c @{\hspace{0.8cm}} c c c @{\hspace{0.8cm}} c c c}
\toprule
\multicolumn{3}{c}{Stash 1} &
\multicolumn{3}{c}{Stash 2} &
\multicolumn{3}{c}{Stash 3} \\
\cmidrule(r){1-3} \cmidrule(r){4-6} \cmidrule(r){7-9}
Val ID & probability & utility &
Val ID & probability & utility &
Val ID & probability & utility \\
\midrule
1 & 0.15 & 0.78 & 1 & 0.18 & 0.78 & 1 & 0.13 & 0.78 \\
2 & 0.12 & 0.76 & 2 & 0.14 & 0.76 & 3 & 0.54 & 0.71 \\
3 & 0.73 & 0.71 & 4 & 0.20 & 0.7  & 4 & 0.14 & 0.7 \\
  &      &      & 5 & 0.48 & 0.63 & 7 & 0.08 & 0.59 \\
  &      &      &   &      &      & 10 & 0.11 & 0.56 \\
\bottomrule
\end{tabular}
}
\end{subtable}

\end{table}

\subsubsection{Convergence and solution strategies}

Convergence of the NSGA-II can be studied by means of the hypervolume indicator \citep{GuerreiroEtAl2021}. We conducted a number of simulations and we found that the hypervolume indicator stagnates after about 100 generations. Stagnation of the hypervolume indicator provides empirical evidence that the approximation of the Pareto front has stabilized. Although this does not prove convergence to the true Pareto front, it is a useful diagnostic for assessing whether additional generations are likely to yield substantial improvements.
%
%
%
Nevertheless, given the large number of validators and the computational complexity of the second objective function \eqref{eq:f2} each iteration could be computationally demanding. Therefore, we considered different strategies to speed up the optimization process while maintaining a high fidelity of the final results. All of them were designed to run for about a minute.

\begin{description}
    \item[Base:] Classical optimization in which all validators are considered, and each candidate solution is fully evaluated at every iteration. Due to high computational costs, the population size was limited to 50, with 200 generations.

    \item[Neural Network:] Optimization based on approximate evaluation, in which a neural network is used to estimate validators’ activation probabilities. These estimates are subsequently employed to compute the expected entropy and expected utility of candidate portfolios. Owing to the substantially lower computational cost of this approximation, we used a population size of 350 and ran the algorithm for 400 generations.

    \item[Limited Input:] The same approach as in the \textit{Neural Network} approach, but applied to a reduced candidate set consisting of the \(n(C+1)\) validators with the highest utility scores, together with at least \(n+2\) validators that were always active in the observed history, so as to ensure that each stash can include at least one such validator.
    
    \item[Hybrid:] The previous strategy, followed by a second optimization stage with a limited number of iterations in which candidate solutions are evaluated exactly rather than through neural network approximations. To contain the computational cost of full evaluations, the population size was reduced to 300, with 300 generations in the first stage and 12 in the second.
\end{description}

In each case, the same initialization method was used. This initialization phase can be seen as a warm start where each stash was assigned a random validator with a $100\%$ probability of being active, along with between 2 and 4 validators randomly selected from the whole dataset. The probability of selection was proportional to the validators' utility. Repetitions were not allowed within a stash but were possible across stashes. 
The mutation operator randomly removed or added a validator for each stash while obeying the minimum and maximum size constraints. The crossover operator, for each stash, randomly copied an entire stash from one of the parents.

\subsection{Finding the most preferred portfolio}
\label{sub:finding}



Selecting a final solution from a Pareto front is a central issue in multi-objective optimization. In \emph{a posteriori} approaches, a set of nondominated solutions is first generated and the \ac{DM} subsequently selects one of them~\citep{EhrgottEtal2026}. This gives the \ac{DM} an overview of the available trade-offs, but may also impose a considerable cognitive burden when the Pareto set is large, when many alternatives are represented only in objective space, or when the objective scales are difficult to interpret. Interactive methods address this difficulty by progressively incorporating preference information during the search or selection process, thereby focusing attention on solutions that are more relevant for the \ac{DM} \citep{Miettinen1999,MiettinenEtAl2008,XinEtAl2018}. Such methods differ in the type of preference information they require~\citep{EhrgottEtal2026}. In reference-point approaches, the \ac{DM} specifies aspiration levels for the objectives and the method searches for nondominated solutions close to these levels, typically through achievement scalarizing functions \citep{Wierzbicki1980}. In stepwise methods such as STEM, the \ac{DM} progressively relaxes some objectives in order to improve others \citep{BenayounEtAl1971}. In trade-off-based approaches, such as the Zionts--Wallenius method, the \ac{DM} provides local preference information, often through responses to trade-off questions, which is then used to guide the search \citep{ZiontsWallenius1976}. Other interactive methods rely on more qualitative or cognitive-friendly forms of feedback. In classification-based methods such as NIMBUS, the \ac{DM} classifies objectives according to whether they should be improved, kept at their current level, or allowed to deteriorate \citep{MiettinenMakela1995}. NAUTILUS-type methods support navigation from inferior objective values toward the Pareto front, so that the \ac{DM} can observe progressive improvements rather than repeatedly accept deterioration in some objectives \citep{MiettinenEtAl2010}. In rule-based approaches, such as the interactive method based on the Dominance-based Rough Set Approach, the \ac{DM} evaluates a sample of solutions and decision rules are induced from these evaluations to focus the search on more promising parts of the Pareto set \citep{GrecoEtAl2008}. These approaches illustrate the variety of preference information that can be used in interactive multi-objective optimization, ranging from aspiration levels and trade-off judgments to objective classifications and examples of satisfactory solutions.

Our setting differs from these classical interactive procedures in two respects. First, the optimization stage produces a finite approximation of the two-objective Pareto front, rather than requiring repeated optimization after each item of preference information. Second, the \ac{DM} is not asked to provide aspiration levels, trade-off rates, objective classifications, or examples of satisfactory and unsatisfactory solutions. Such information could be difficult to elicit because expected utility and expected entropy are abstract scales, while nominators are more naturally able to reason about concrete nomination portfolios. 

We therefore use a simple binary search procedure to select a final solution, tailored to the ordered structure of a two-objective Pareto front. Let the Pareto-optimal solutions be ordered as $\mathbf{p}=(p_1,\ldots,p_s)$ from left to right in the objective space, where $p_1$ maximizes the second objective and $p_s$ maximizes the first objective. The procedure starts from the middle solution, namely $p^{(0)}:=p_{\lceil s/2\rceil}$, and presents the corresponding portfolio allocation to the \ac{DM} in the decision space.

At each step, the \ac{DM} is shown the current portfolio and is asked whether the search should move toward higher expected utility or toward higher diversification. The next candidate is then selected as the middle solution within the subset indicated by the \ac{DM} (i.e., improving the selected objective while deteriorating the other), and its corresponding portfolio allocation is presented for evaluation. In this way, the procedure keeps the interaction close to the actual nomination problem while exploiting the ordered structure of the Pareto front to reduce the number of required comparisons.

The process is repeated until either an extreme solution is reached or the \ac{DM} is satisfied with the currently displayed portfolio. Since each step halves the set of candidate solutions, the maximum number of iterations required is $\lceil \log_2 s \rceil$. For example, if the Pareto front contains $s=64$ solutions, at most six questions are needed. In practice, the procedure may terminate earlier whenever the \ac{DM} finds the current solution satisfactory.

The main phases of the proposed approach are summarized in Figure~\ref{fig:framework}. Nominator preferences and blockchain data are first used to estimate validator utilities and allocation probabilities. These components feed the bi-objective optimization model, which produces a personalized Pareto front of efficient portfolio collections. The nominator then navigates this front through a binary-search procedure to select the final recommendation.

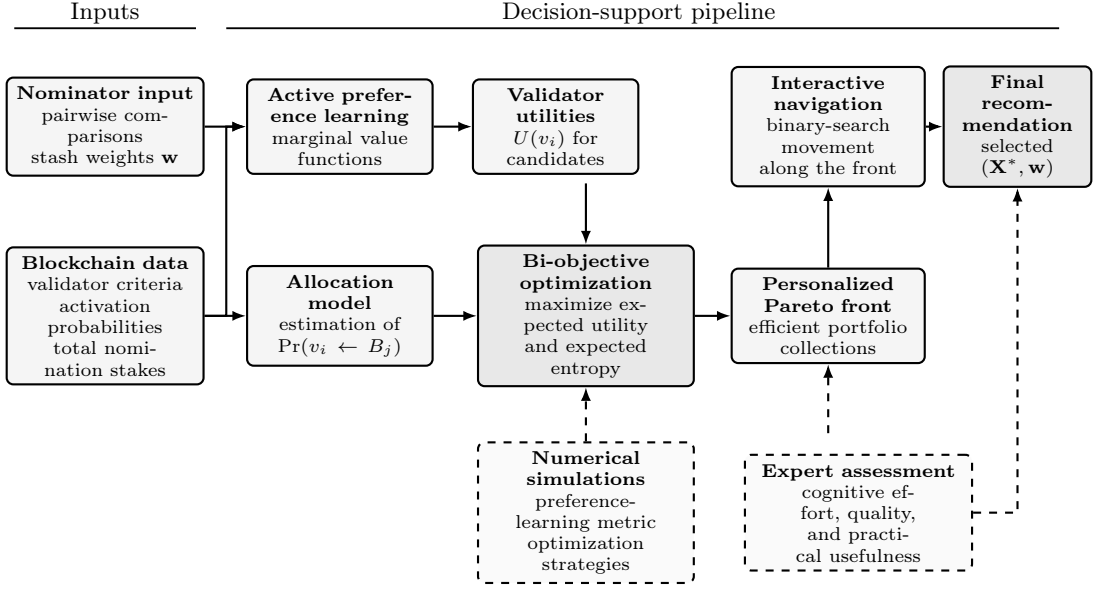
\begin{figure}[t]
\centering
\begin{tikzpicture}[
    x=1cm,
    y=1cm,
    font=\scriptsize,
    >=Latex,
    box/.style={
        draw=black,
        rounded corners=2.5pt,
        line width=0.8pt,
        align=center,
        fill=gray!8
    },
    optbox/.style={
        draw=black,
        rounded corners=2.5pt,
        line width=0.8pt,
        align=center,
        fill=gray!18
    },
    valbox/.style={
        draw=black,
        rounded corners=2.5pt,
        line width=0.8pt,
        dashed,
        align=center,
        fill=gray!4
    },
    arr/.style={
        -{Latex[length=1.8mm]},
        line width=0.8pt
    },
    darr/.style={
        -{Latex[length=1.8mm]},
        line width=0.8pt,
        dashed
    }
]

\node[font=\small] at (1.4,6.45) {Inputs};
\draw[line width=0.7pt] (0.2,6.25) -- (2.6,6.25);

\node[font=\small] at (8.45,6.45) {Decision-support pipeline};
\draw[line width=0.7pt] (3.0,6.25) -- (14.0,6.25);


\node[box, minimum width=2.6cm, minimum height=0.95cm, text width=2.35cm] (dm)
at (1.4,4.95)
{\textbf{Nominator input}\\
pairwise comparisons\\
stash weights $\mathbf{w}$};

\node[box, minimum width=2.6cm, minimum height=1.05cm, text width=2.35cm] (data)
at (1.4,2.45)
{\textbf{Blockchain data}\\
validator criteria\\
activation probabilities\\
total nomination stakes};

\node[box, minimum width=2.45cm, minimum height=0.95cm, text width=2.2cm] (learning)
at (4.5,4.95)
{\textbf{Active preference learning}\\
marginal value functions};

\node[box, minimum width=2.45cm, minimum height=1.05cm, text width=2.2cm] (allocation)
at (4.5,2.45)
{\textbf{Allocation model}\\
estimation of\\
$\Pr(v_i \leftarrow B_j)$};

\node[box, minimum width=2.15cm, minimum height=0.95cm, text width=1.95cm] (utility)
at (7.35,4.95)
{\textbf{Validator utilities}\\
$U(v_i)$ for candidates};

\node[optbox, minimum width=2.85cm, minimum height=1.35cm, text width=2.55cm] (optimization)
at (7.75,2.45)
{\textbf{Bi-objective optimization}\\
maximize expected utility\\
and expected entropy};

\node[box, minimum width=2.55cm, minimum height=1.05cm, text width=2.25cm] (pareto)
at (10.95,2.45)
{\textbf{Personalized Pareto front}\\
efficient portfolio\\
collections};

\node[box, minimum width=2.55cm, minimum height=0.95cm, text width=2.25cm] (navigation)
at (10.95,4.95)
{\textbf{Interactive navigation}\\
binary-search movement\\
along the front};

\node[box, fill=gray!18, minimum width=1.95cm, minimum height=0.95cm, text width=1.7cm] (final)
at (13.45,4.95)
{\textbf{Final recommendation}\\
selected\\
$(\mathbf{X}^{*},\mathbf{w})$};

\node[valbox, minimum width=2.85cm, minimum height=0.9cm, text width=2.55cm] (sim)
at (7.75,-0.15)
{\textbf{Numerical simulations}\\
preference-learning metric\\
optimization strategies};

\node[valbox, minimum width=3.0cm, minimum height=0.9cm, text width=2.7cm] (expert)
at (11.35,-0.15)
{\textbf{Expert assessment}\\
cognitive effort, quality,\\
and practical usefulness};


\coordinate (expertParetoStart) at (10.95,0.90);


\draw[arr] (dm.east) -- (learning.west);
\draw[arr] (data.east) -- (allocation.west);
\draw[arr] (learning.east) -- (utility.west);
\draw[arr] (allocation.east) -- (optimization.west);   
\draw[arr] (optimization.east) -- (pareto.west);       
\draw[arr] (navigation.east) -- (final.west);

\draw[arr] (data.east) -- (3.0,2.45) -- (3.0,4.95) -- (learning.west);

\draw[arr] (7.75,4.2) -- (optimization.north);

\draw[arr] (pareto.north) -- (navigation.south);

\draw[darr] (sim.north) -- (optimization.south);

\draw[darr] (expertParetoStart) -- (pareto.south);

\draw[darr] (expert.east) -- (13.45,-0.15) -- (final.south);

\end{tikzpicture}
\caption{Main phases of the proposed decision support framework (dashed boxes indicate the experimental validation components).}
\label{fig:framework}
\end{figure}

\section{Experimental results and validation}
\label{sec:experiments}
{We followed two main directions to validate our approach. The first one, by means of numerical simulations, aimed at verifying the speed and the reliability of the optimization problem, whereas the second one involved real \acp{DM} with the goal of studying the empirical perception and acceptability of the decision support system.}

\subsection{Simulations}

To increase precision and speed of convergence, as described in the previous section, we had to customize parts of the algorithmic procedure. To validate the novel approaches, also compared to the classical approaches, we carried out a series of experiments with simulated \acp{DM}.

\begin{figure}[htbp]
    \centering

    \begin{subfigure}{0.48\textwidth}
        \centering
        \includegraphics[width=\linewidth]{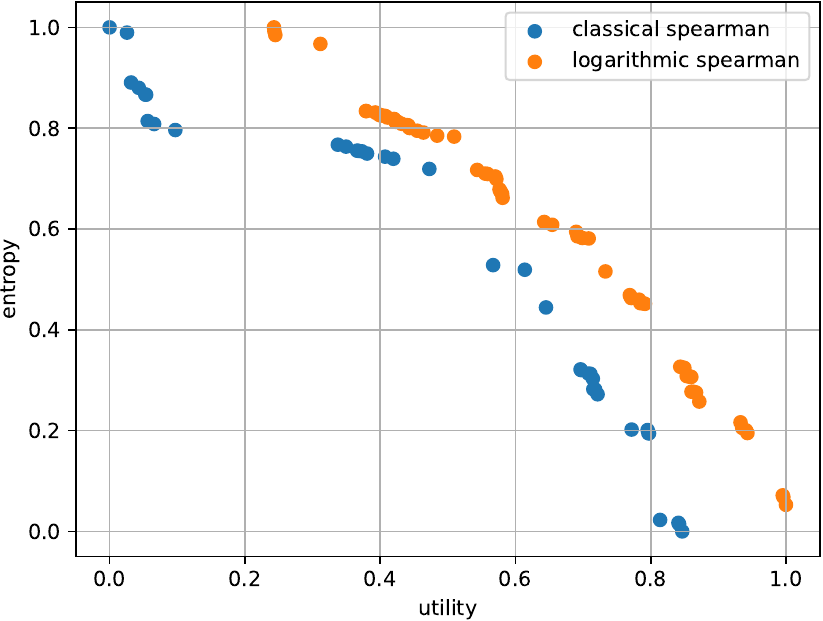}
        \caption{Comparison of solutions obtained for different metrics comparing rankings.}
        \label{fig:selection}
    \end{subfigure}\hfill
    \begin{subfigure}{0.48\textwidth}
        \centering
        \includegraphics[width=\linewidth]{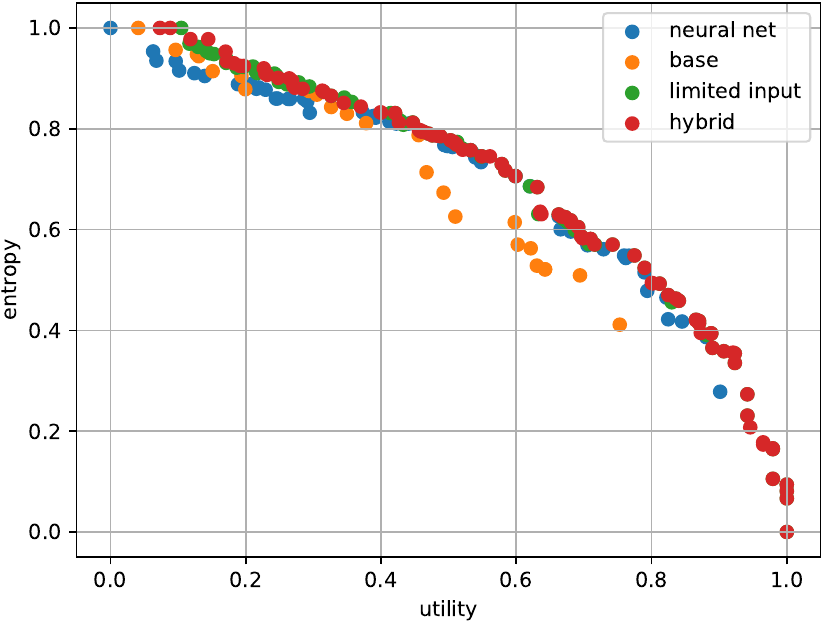}
        \caption{Example of Pareto fronts obtained for all four optimization strategies. The performances are scaled to fit into range [0-1].}
        \label{fig:comparison}
    \end{subfigure}

    \vspace{0.5em}

    \begin{subfigure}{0.48\textwidth}
        \centering
        \includegraphics[width=\linewidth]{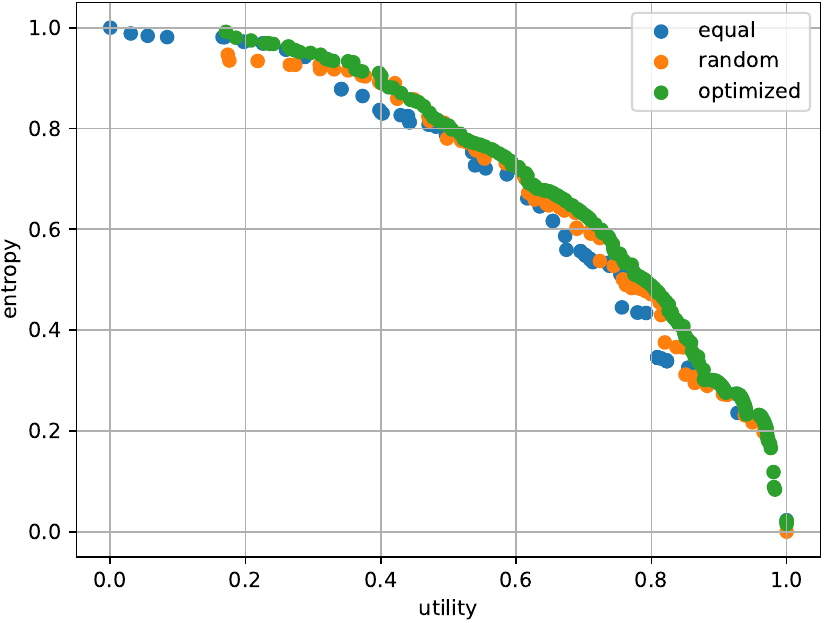} 
        \caption{Example of Pareto fronts obtained for all three considered weights strategies. The performances are scaled to fit into range [0-1].}
        \label{fig:weights}
    \end{subfigure}\hfill
    \begin{subfigure}{0.48\textwidth}
     \centering
\includegraphics[width=\linewidth]{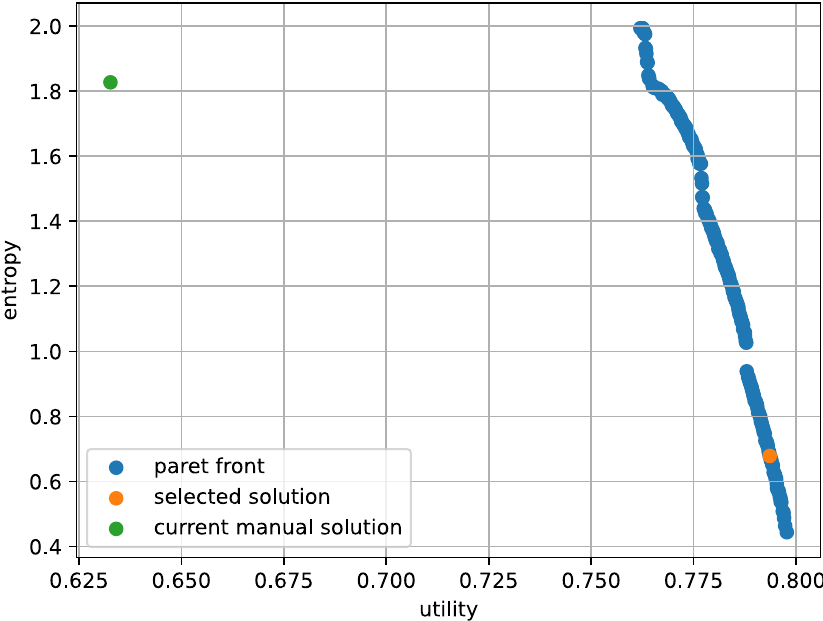} 
    \caption{Pareto front obtained for a single \ac{DM} together with solution selected by them and a point representing their current manually selected portfolios}
    \label{fig:compr}
    \end{subfigure}

    \caption{Result obtained with different strategies.}
    \label{fig:all}
\end{figure}

\subsubsection{Preference learning}

The first phase of our algorithmic procedure is based on the preference learning procedure proposed by \citet{GehrleinEtAl2023}, where the optimal sequence of questions that had to be asked to the \acp{DM} was determined by sequentially minimizing the size of the largest set of compatible value models left after the \acp{DM} provided new preferential information. This size was estimated by the maximum Spearman distance between rankings of alternatives that could have been obtained with a compatible model. Since in our case only the best validators actually matter, we tested a new approach where the ranks of alternatives were scaled using a logarithmic transformation so that discrepancies between high-ranked alternatives matter more than those on lower-ranked alternatives. In practice, a new method of estimating the information gain by a Pearson correlation of logarithmically transformed ranks was compared with classical Spearman rank correlation as follows:
\begin{enumerate}
    \item Create a random \ac{DM} with randomly selected utility functions;
    \item Use the \ac{DM} in the preference learning procedure with both metrics separately;
    \item Apply the portfolio optimization method for both obtained utility functions;
    \item Recalculate the objectives based on the true utility functions of the \ac{DM};
    \item Compare the resulting hypervolumes.
\end{enumerate}

To objectively compare different scenarios, the results were first standardized using a min-max scaling on each objective independently. The results obtained using the proposed metric outperformed those obtained with the classical Spearman rank correlation, achieving, on average, a 38\% higher standardized hypervolume and outperforming the classical metric in 65\% of the cases.
An example of the results obtained with both metrics is presented in Figure~\ref{fig:selection}.


\subsubsection{Bi-objective optimization}

We evaluated the four optimization strategies presented in Section~\ref{sub:bi-objective} across multiple test cases, with a time limit of one minute per run. In each run, a fixed set of real validators was used, while their utility values were randomly assigned. The number of stashes was randomly selected from the range 2 to 4, and stash weights were generated independently for each run. The hybrid scenario achieved the best overall performance, statistically significantly outperforming all remaining approaches with paired t-tests returning $p < 0.05$; therefore, it was used in the later experiments. A comparison of the Pareto fronts for a representative run is shown in Figure~\ref{fig:comparison} and the detailed results are presented in Table~\ref{tab:scenarios}.

\begin{table}[ht]
\centering
\begin{tabular}{lcc}
\toprule
 Method & Avg hypervolume & Avg rank \\
\midrule
\multicolumn{3}{c}{\textbf{(a) Optimization scenario experiment}} \\
     Base    & 0.555 & 4 \\
     Neural Network    & 0.725 & 2.87\\
     Limited input & 0.771 & 1.65\\
     Hybrid & 0.773 & 1.48\\
\midrule
\multicolumn{3}{c}{\textbf{(b) Weights experiment}} \\
     Random    & 0.56 & 2.62 \\
     Equal    & 0.639 & 2.08\\
     Optimized & 0.652 & 1.31\\
\bottomrule
\end{tabular}
\caption{Mean hypervolume and rank obtained for four different optimization scenarios in an experiment with randomized data and for the optimization with 3 different approaches to weights associated with stashes}
\label{tab:scenarios}
\end{table}




\subsubsection{Adjustable weights}
Due to limitations of the system, the weights stored in the vector $\mathbf{w}$, representing the funds associated with each stash, were kept constant. Since the weight distribution plays an important role in the calculation of both objectives, an additional experiment was conducted to evaluate the results when the weights were allowed to change during the optimization process. In a real-world application, modifying the weights would require transferring funds between stashes. A comparison of the Pareto fronts for a representative run is shown in Figure~\ref{fig:weights}, while the detailed results are presented in Table~\ref{tab:scenarios}.

It is worth noting that both extreme solutions, using the same validators in each stash to maximize utility, and enforcing no overlap between stashes to maximize entropy, achieve the best performance under an equal distribution of funds. Nevertheless, allowing the weights to vary enables the discovery of trade-offs between the two objectives that cannot be achieved when the weights are fixed.


Three different scenarios were considered and tested over multiple runs:
\begin{description}
    \item[Random weights:] Weights were selected randomly from a uniform distribution. In each optimization run they were selected again.
    \item[Equal weights:] Each stash was associated with the same weight.
    \item[Optimized weights:] Weights were adjusted during the optimization process.
\end{description}

\subsubsection{Improvement by adding a stash}

Each additional stash allows higher entropy to be reached and provides more flexibility in finding a trade-off between the two criteria. To check the impact of adding an additional stash, an experiment was performed. Given the same set of validators, the optimization task was performed for $b$ and $b+1$ stashes, with equal weights across stashes. Then, we studied the average improvement of the hypervolume indicators when one more stash was added. The results show that the marginal benefit of adding one more stash decreases as the number of existing stashes increases. The improvement exceeds $100\%$ when switching from 2 to 3 stashes and falls below $20\%$ when the number of stashes is 5 or more.


\subsection{Expert evaluation with real Decision Makers}
To evaluate the proposed approach in practice, we conducted an empirical evaluation of our proposed algorithm with real nominators from the Polkadot ecosystem. The study was carried out online between March and April 2026, with five participants who actively stake on the network and face the validator selection problem with multiple stashes. The study followed a structured, multi-stage protocol, which we describe in turn before reporting the results. We emphasized that we do not collect any personalized data and that participation is anonymous.

\begin{description}
\item[Initialization.] Participants received online instructions about the upcoming study and the prerequisites to participate. These included that they were currently actively staking with multiple stashes. We also asked them to note the current distribution of funds across their stashes, an input required by our algorithm for the optimization. Within the instructions, they found a link to start the algorithm and the final questionnaire.

\item[Utility Elicitation.] Participants' preferences over validator criteria were elicited following the active preference learning framework of \citet{GehrleinEtAl2023}. In this stage, participants were presented with pairs of validators characterized by their criteria values and asked to indicate which they preferred. By working through a sequence of such pairwise comparisons, the procedure recovers marginal utility functions over all criteria, yielding a utility score $U^{k}(v_i)$ for every validator $v_i \in V$ and each nominator~$k$.

\item[Pareto Front Construction.] The bi-objective optimization problem \eqref{eq:bi-objective} was solved individually for each nominator using their elicited utility function $U^{k}$, producing a personalized set of non-dominated portfolios as illustrated in Figure~\ref{fig:Pareto}.

\item[Portfolio Selection.] Participants were then asked to identify their preferred solution from the Pareto front by navigating the trade-off between expected utility and diversification. They employed the interactive procedure described in Section~\ref{sub:finding}, iterating until converging on a Pareto-efficient portfolio with which they were satisfied.

\item[Final Recommendation.] Participants were presented side by side with their current nomination portfolio and the algorithm's recommended portfolio, and asked to compare the two carefully.

\item[Questionnaire.] Finally, participants completed a structured questionnaire about their experiences.
\end{description}

Upon completion of the online interaction, the questionnaire was designed to assess three dimensions of the proposed decision support system: (i)~the cognitive effort required at each stage, (ii)~the perceived quality of the recommendation, and (iii)~the overall acceptance of the approach relative to the participants' current validator selection practices.

The sample, though small, consists of highly experienced nominators and thereby constitutes an expert evaluation of our algorithm. Four out of five participants (80\%) reported staking on Polkadot for more than three years, with the remaining participant having between one and three years of experience. This level of familiarity with the staking process is valuable for the validity of the evaluation, as these \acp{DM} are well-positioned to judge the merits and limitations of the proposed system against their established practices.

\subsubsection{Results}

Table~\ref{tab:summary} summarizes the key findings across all evaluation dimensions. The results reveal a clear and informative pattern across the two main stages of the interaction. The first stage, in which participants answer pairwise comparison questions to elicit their preferences over validator attributes, was unanimously perceived as easy. All five participants rated this step as either ``easy'' or ``very easy.'' This finding is consistent with the results of the larger-scale experiment reported in \cite{GehrleinEtAl2023} and confirms that the active preference learning procedure, which decomposes a complex multi-attribute evaluation into a sequence of binary choices, remains cognitively manageable even when the elicited preferences are subsequently used in a more complex optimization context.

\begin{table}[htbp]
\centering \footnotesize
\caption{Summary of the empirical evaluation results.}
\label{tab:summary}
\begin{tabular}{ll}
\toprule
\textbf{Metric} & \textbf{Result} \\
\midrule
Number of participants $n$ & 5 \\
Pairwise comparison difficulty & 100\% rated Easy or Very Easy \\
Portfolio adjustment difficulty & 80\% rated Difficult \\
Optimal portfolio rating & $M = 5.4$, $Mdn = 6$ (scale 1--7) \\
Would follow recommendation & 3\,/\,5 (60\%) \\
Process is easier & 3\,/\,5 agree or strongly agree \\
Process is more convenient & 3\,/\,5 agree or strongly agree \\
Process is helpful & 5\,/\,5 agree or strongly agree \\
Staking experience & 80\% stake 3+ years \\
\bottomrule
\end{tabular}
\end{table}

The second stage, in which participants adjusted the recommended portfolio by navigating the Pareto front of non-dominated solutions, was rated as ``difficult'' by four out of five participants (80\%). This contrast is notable and suggests that, while the preference elicitation component of the system is well-suited to the capabilities of the target users, the portfolio adjustment step --- where nominators must reason about the trade-off between expected utility and diversification (entropy) --- introduces a substantially higher cognitive load. This result is to be expected because the underlying task is inherently complex. But despite this, 3/5 participants agreed that the proposed process is easier than whichever alternative the \acp{DM} currently use. In other words, the underlying task is very complex, but although still challenging, our approach eased it.

Despite the challenge posed by the portfolio adjustment stage, participants evaluated the quality of the final recommendation favorably. On a scale from 1 to 7, the mean rating was 5.4 ($SD = 0.89$) with a median of 6. Three participants gave a score of 6, one a score of 5, and one a score of 4. These ratings indicate that the optimization procedure produces portfolios that nominators consider to be of good quality, even if the process of arriving at them is perceived as demanding.

When asked whether they would follow the algorithm's recommendation, three out of five participants (60\%) responded affirmatively. The two participants who indicated they would not follow the recommendation nonetheless provided ratings that shed important light on the nature of their reluctance. One of these participants rated the portfolio quality at 6 out of 7 and explicitly acknowledged the quality of the suggested validators in their open-ended comments. Specifically, this participant noted: ``The algorithm suggested 3 validators for stash 1 and 4 validators for stash 2. Although the quality of the validators is good, I would still be inclined to nominate more than 10 validators per stash to ensure I receive my rewards, even though I do not monitor my staking rewards for several weeks. I might include the suggested validators in the set, but I will be adding more according to my preferences.'' This response reveals that the participant's reluctance to follow the recommendation was not driven by dissatisfaction with the quality of the selected validators, but rather by a preference for larger nomination sets as a risk mitigation strategy. This participant explicitly stated they would include the suggested validators in their nomination, but wished to supplement them with additional choices. This insight is particularly relevant to the practical deployment of the system, as it suggests that the tool's output could serve as a high-quality starting point that nominators can augment according to their individual risk preferences, rather than a prescriptive final selection.

The comparative assessment of the proposed approach against participants' current validator selection practices produced a mixed but overall encouraging picture. Three out of five participants agreed or strongly agreed that the process would be easier and more convenient than their current approach, while two disagreed. Importantly, the strongest consensus emerged on the question of helpfulness: all five participants (100\%) agreed or strongly agreed that the process would be ``really helpful.'' This unanimous endorsement of the system's usefulness, even among those who found it neither easier nor more convenient than manual selection, suggests that the proposed approach offers value beyond mere convenience. Participants recognized that the system delivers benefits that their current ad hoc methods cannot match, even though interacting with the system requires effort.

The cross-tabulation of responses further supports this interpretation. The two participants who would not follow the recommendation also disagreed that the process was easier or more convenient, suggesting a coherent stance: these are experienced nominators with established workflows who see the tool as informative but not as a replacement for their current approach. Conversely, the three participants who would follow the recommendation consistently agreed or strongly agreed on ease and convenience, indicating that for a subset of users, the system already provides a satisfactory end-to-end solution.

\subsubsection{Qualitative insights}

The open-ended comments provided by participants offer additional depth to the quantitative findings. Beyond the detailed comment on portfolio size discussed above, one participant highlighted that a key factor in their personal validator selection process (whether they personally know the operator of the validator) is not captured by the attribute-based model. This observation underscores a well-known limitation of formal decision support systems: they operate on quantifiable and observable criteria, while a \ac{DM} may incorporate soft, relational, or reputational information that is inherently difficult to formalize. In the context of blockchain governance, trust relationships and personal knowledge of validator operators constitute an important dimension of the selection problem that complements the quantitative attributes considered in the model. Future iterations of the system could address this by allowing nominators to manually include or exclude validators based on such qualitative considerations, effectively using the algorithm's output as one input among several in a hybrid decision process.

\subsubsection{Discussion}

Taken together, the empirical results paint a coherent picture of a system that is perceived as helpful and capable of generating high-quality recommendations, but inherits the underlying complexity of the problem it tries to solve. The preference learning component on individual validator properties~\cite{GehrleinEtAl2023} continues to perform well in terms of cognitive effort. The extension to the portfolio optimization setting introduces additional complexity that is reflected in the participants' difficulty ratings, but not in a diminished appreciation of the output quality or the overall value of the approach.

The finding that even participants who would not adopt the recommendation in full recognized the quality of the suggested validators is encouraging from a practical standpoint. It suggests that the system can serve different user profiles: for some nominators, the recommended portfolio may be adopted directly; for others, it may function as a curated shortlist to be integrated into a broader, personally informed selection strategy. This flexibility aligns well with the insight from \cite{dietvorst2018overcoming} that \acp{DM} are more willing to use algorithmic aids when they retain the ability to modify the output, even if only slightly.

We acknowledge the limitations of this empirical evaluation. The sample size of five participants, while sufficient to identify salient patterns and collect rich qualitative feedback, does not permit formal statistical inference. The results should therefore be interpreted as indicative rather than conclusive. A larger-scale deployment of the system, ideally integrated into the existing staking infrastructure, would enable a more rigorous evaluation and allow for statistical comparisons across different user segments and staking strategies.

\subsubsection{Comparison}
One participant shared their complete nomination data, including the proportion of funds associated with each stash and the validators nominated in each case. Combined with the preference information provided in the form of pairwise comparisons, this made it possible to compare a solution manually defined by the \ac{DM} with the result of the optimization process. The obtained Pareto front, together with a point representing the current portfolio of the \ac{DM}, is presented in Figure~\ref{fig:compr}. The solution manually defined by the \ac{DM} is characterized by high entropy due to the large number of validators involved, while its average utility ($0.63$) is significantly lower than the utilities of the solutions forming the Pareto front.


\section{Conclusions}
\label{sec:conclusions}

This paper addressed the problem of supporting validator nomination in proof-of-stake blockchains when nominators operate through multiple stashes. In such settings, the decision problem goes beyond selecting a set of individually attractive validators. A nominator must construct a collection of nomination portfolios whose final consequences depend on an election and allocation mechanism that is only partially controlled by the nominator. This makes the problem structurally different from standard portfolio selection models: the relevant exposure is induced by a stochastic allocation process, the same validator may appear in several stashes, and the quality of a solution depends jointly on expected validator performance and diversification across the final allocation.

From a methodological perspective, we formulated this task as a bi-objective portfolio collection optimization problem. The first objective captures the quality and profitability of the portfolio collection, while the other reflects the diversification and risk mitigation across validators and stashes. This formulation provides a transparent representation of the trade-off faced by nominators: concentrating nominations on high-utility validators may improve expected quality, while distributing exposure more broadly can reduce allocation risk. The model also incorporates an explicit probabilistic representation of validator allocation, based on activation probabilities and the relative likelihood of validators being assigned from a~stash.

The proposed decision support framework combines preference learning, multi-objective optimization, and interactive selection. Validator utilities are inferred using a dedicated active preference learning procedure. The resulting optimization problem is solved using a~multi-objective evolutionary algorithm, and several computational strategies are examined to obtain high-quality Pareto fronts within a practically acceptable time. To support the final choice, we introduced a simple binary-search navigation procedure over the ordered two-objective Pareto front.

The experimental results support the relevance of the proposed approach. The simulations indicate that the modified criterion on active preference elicitation improves the quality of the optimized portfolios relative to the classical rank-based criterion. The comparison of optimization strategies shows that approximation and candidate-reduction mechanisms can substantially improve computational performance, with the hybrid strategy achieving the best overall results among the tested variants. These findings are important because the second objective is computationally demanding, and an operational decision support system must produce recommendations within a short interaction time.

The empirical evaluation with experienced Polkadot nominators provides complementary evidence of the approach's practical value. Although the sample is necessarily limited, the participants were active and knowledgeable users facing the problem addressed in this paper. The preference elicitation stage was perceived as easy, confirming that pairwise comparisons over validator attributes remain a cognitively manageable way to elicit preferences. The portfolio adjustment stage was perceived as more difficult, which reflects the inherent complexity of reasoning about expected utility and diversification. Nevertheless, the final recommendations were well received, and all participants found the process helpful. These results suggest that the proposed system can provide value both as an end-to-end recommendation tool for some nominators and as a structured source of high-quality portfolio suggestions for users who wish to adjust further the recommendations based on their own experience, trust relationships, or risk attitudes.

The study also opens several directions for future research. First, the empirical validation should be extended to a larger group of nominators and, ideally, to a deployment integrated with existing staking tools. This would enable analysis of adoption, long-term usage, and realized staking outcomes. Second, future versions of the model could allow nominators to impose additional constraints, such as mandatory inclusion or exclusion of specific validators, minimum portfolio sizes, or requirements reflecting personal trust in validator operators. Third, the model could be generalized to alternative blockchain protocols and election rules, including settings which allow multiple active validators from a single stash. Finally, the proposed framework may also be useful from a governance perspective. Simulating how optimized nomination portfolios change under different protocol rules or allocation mechanisms could support the analysis of incentives and decentralization in proof-of-stake ecosystems.

\section*{Acknowledgments}
\noindent G. Miebs and A. Mielniczuk were supported by the Polish Ministry of Science and Higher Education, grant no. 0311/SBAD/0760. M. Kadzi{\'n}ski acknowledges the support of the Polish Ministry of Science and Higher Education, grant no. 0311/SBAD/0760.

\bibliographystyle{apalike}
\bibliography{biblio}

@article{GehrleinEtAl2023,
  title={An active preference learning approach to aid the selection of validators in blockchain environments},
  author={Gehrlein, Jonas and Miebs, Grzegorz and Brunelli, Matteo and Kadzi{\'n}ski, Mi{\l}osz},
  journal={Omega},
  volume={118},
  pages={102869},
  year={2023},
  publisher={Elsevier}
}

@article{MasmoudiAbdelaziz2018,
  author  = {Masmoudi, Meryem and Ben Abdelaziz, Fouad},
  title   = {{Portfolio selection problem: A review of deterministic and stochastic multiple objective programming models}},
  journal = {Annals of Operations Research},
  volume  = {267},
  number  = {1--2},
  pages   = {335--352},
  year    = {2018},
  doi     = {10.1007/s10479-017-2466-7}
}

@book{SaloKeislerMorton2011,
  editor    = {Salo, Ahti and Keisler, Jeffrey and Morton, Alec},
  title     = {{Portfolio Decision Analysis: Improved Methods for Resource Allocation}},
  publisher = {Springer},
  address   = {New York},
  series    = {International Series in Operations Research \& Management Science},
  volume    = {162},
  year      = {2011},
  doi       = {10.1007/978-1-4419-9943-6}
}

@article{GutjahrEtal2008,
  author  = {Gutjahr, Walter J. and Katzensteiner, Stefan and Reiter, Philipp and Stummer, Christian and Denk, Martin},
  title   = {Competence-driven project portfolio selection, scheduling and staff assignment},
  journal = {Central European Journal of Operations Research},
  volume  = {16},
  number  = {3},
  pages   = {281--306},
  year    = {2008},
  doi     = {10.1007/s10100-008-0057-z}
}

@article{GutjahrReiter2010,
  author  = {Gutjahr, Walter J. and Reiter, Philipp},
  title   = {Bi-objective project portfolio selection and staff assignment under uncertainty},
  journal = {Optimization},
  volume  = {59},
  number  = {3},
  pages   = {417--445},
  year    = {2010},
  doi     = {10.1080/02331931003700699}
}

@article{GutjahrEtal2010,
  author  = {Gutjahr, Walter J. and Katzensteiner, Stefan and Reiter, Philipp and Stummer, Christian and Denk, Martin},
  title   = {Multi-objective decision analysis for competence-oriented project portfolio selection},
  journal = {European Journal of Operational Research},
  volume  = {205},
  number  = {3},
  pages   = {670--679},
  year    = {2010},
  doi     = {10.1016/j.ejor.2010.01.041}
}

@article{NoroDias2023,
  author  = {Noro, Jorge and Dias, Lu{\'i}s C.},
  title   = {Project portfolio management considering the commitment of agents: A bi-objective model applied to administrative services},
  journal = {Journal of the Operational Research Society},
  volume  = {74},
  number  = {4},
  pages   = {1049--1062},
  year    = {2023},
  doi     = {10.1080/01605682.2022.2056530}
}

@article{KalayciEtAl2019,
  author  = {Kalayci, Can B. and Ertenlice, Okkes and Akbay, Mehmet Anil},
  title   = {{A comprehensive review of deterministic models and applications for mean-variance portfolio optimization}},
  journal = {Expert Systems with Applications},
  volume  = {125},
  pages   = {345--368},
  year    = {2019},
  doi     = {10.1016/j.eswa.2019.02.011}
}

@article{MetaxiotisLiagkouras2012,
  author  = {Metaxiotis, Konstantinos and Liagkouras, Konstantinos},
  title   = {{Multiobjective Evolutionary Algorithms for Portfolio Management: A comprehensive literature review}},
  journal = {Expert Systems with Applications},
  volume  = {39},
  number  = {14},
  pages   = {11685--11698},
  year    = {2012},
  doi     = {10.1016/j.eswa.2012.04.053}
}

@article{DebEtAl2002,
  title={A fast and elitist multiobjective genetic algorithm: {NSGA-II}},
  author={Deb, Kalyanmoy and Pratap, Amrit and Agarwal, Sameer and Meyarivan, TAMT},
  journal={IEEE Transactions on Evolutionary Computation},
  volume={6},
  number={2},
  pages={182--197},
  year={2002},
  publisher={IEEE}
}

@article{SaloEtAl2024,
   title={Fifty years of portfolio optimization},
  author={Salo, Ahti and Doumpos, Michalis and Liesi{\"o}, Juuso and Zopounidis, Constantin},
  journal={European Journal of Operational Research},
  volume={318},
  number={1},
  pages={1--18},
  year={2024},
  publisher={Elsevier}
}

@article{Saleh2021,
  title={Blockchain without waste: Proof-of-stake},
  author={Saleh, Fahad},
  journal={The Review of Financial Studies},
  volume={34},
  number={3},
  pages={1156--1190},
  year={2021},
  publisher={Oxford University Press}
}

@article{BentovEtAl2014,
  title={Proof of activity: Extending bitcoin's proof of work via proof of stake},
  author={Bentov, Iddo and Lee, Charles and Mizrahi, Alex and Rosenfeld, Meni},
  journal={Performance Evaluation Review},
  volume={42},
  number={3},
  pages={34--37},
  year={2014},
  publisher={ACM New York, NY, USA}
}

@article{BlankDeb2020,
  title={Pymoo: Multi-objective optimization in python},
  author={Blank, Julian and Deb, Kalyanmoy},
  journal={IEEE Access},
  volume={8},
  pages={89497--89509},
  year={2020},
  publisher={IEEE}
}

@article{BarbatiGrecoFigueira2025,
  title={A Multi-Objective Portfolio of Portfolios Problem with Qualitative Performance Assessments},
  author={Barbati, Maria and Greco, Salvatore and Figueira, Jos{\'e} Rui},
  journal={arXiv preprint arXiv:2503.02373},
  year={2025}
}

@article{BrillEtAl2024,
  title={Phragm{\'e}n’s voting methods and justified representation},
  author={Brill, Markus and Freeman, Rupert and Janson, Svante and Lackner, Martin},
  journal={Mathematical Programming},
  volume={203},
  number={1},
  pages={47--76},
  year={2024},
  publisher={Springer}
}

@article{VanHaaren2023,
  title={The seven capital sins in the governance of blockchain ecosystems},
  author={Van Haaren-van Duijn, Birgitte and Roca, Jaime Bonn{\'\i}n and Romme, A Georges L and Weggeman, Mathieu},
  journal={IEEE Engineering Management Review},
  volume={51},
  number={3},
  pages={13--17},
  year={2023},
  publisher={IEEE}
}

@article{EhrgottEtal2026,
  title={{Fifty years of multi-objective optimization and decision-making: From mathematical programming to evolutionary computation}},
  author={Ehrgott, Matthias and K{\"o}ksalan, Murat and Kadzi{\'n}ski, Mi{\l}osz and Deb, Kalyanmoy},
  journal={European Journal of Operational Research},
 volume = {330},
 number = {1},
 pages = {1-25},
 year = {2026},
 issn = {0377-2217},
 doi = {https://doi.org/10.1016/j.ejor.2025.06.012}, 
 publisher={Elsevier}
}

@article{LiesioEtAl2021,
  title={Portfolio decision analysis: Recent developments and future prospects},
  author={Liesi{\"o}, Juuso and Salo, Ahti and Keisler, Jeffrey M and Morton, Alec},
  journal={European Journal of Operational Research},
  volume={293},
  number={3},
  pages={811--825},
  year={2021},
  publisher={Elsevier}
}

@article{HuangEtAl2024,
  title={An overview of Web3 technology: Infrastructure, applications, and popularity},
  author={Huang, Renke and Chen, Jiachi and Wang, Yanlin and Bi, Tingting and Nie, Liming and Zheng, Zibin},
  journal={Blockchain: Research and Applications},
  volume={5},
  number={1},
  pages={100173},
  year={2024},
  publisher={Elsevier}
}

@article{GorccunEtAl2023,
  title={The blockchain technology selection in the logistics industry using a novel {MCDM} framework based on {Fermatean} fuzzy sets and {Dombi} aggregation},
  author={G{\"o}r{\c{c}}{\"u}n, {\"O}mer Faruk and Pamucar, Dragan and Biswas, Sanjib},
  journal={Information Sciences},
  volume={635},
  pages={345--374},
  year={2023},
  publisher={Elsevier}
}

@article{KrishankumarEtAl2024,
  title={Selection of a viable blockchain service provider for data management within the internet of medical things: An {MCDM} approach to {Indian} healthcare},
  author={Krishankumar, Raghunathan and Dhruva, Sundararajan and Ravichandran, Kattur S and Kar, Samarjit},
  journal={Information Sciences},
  volume={657},
  pages={119890},
  year={2024},
  publisher={Elsevier}
}

@article{BuyukozkanEtAl2021,
  title={A decision-making framework for evaluating appropriate business blockchain platforms using multiple preference formats and {VIKOR}},
  author={B{\"u}y{\"u}k{\"o}zkan, G{\"u}l{\c{c}}in and T{\"u}fekci, Gizem},
  journal={Information Sciences},
  volume={571},
  pages={337--357},
  year={2021},
  publisher={Elsevier}
}

@inproceedings{10.1145/3479722.3480988,
author = {Cevallos, Alfonso and Stewart, Alistair},
title = {A verifiably secure and proportional committee election rule},
year = {2021},
isbn = {9781450390828},
publisher = {Association for Computing Machinery},
address = {New York, NY, USA},
url = {https://doi.org/10.1145/3479722.3480988},
doi = {10.1145/3479722.3480988},
booktitle = {Proceedings of the 3rd ACM Conference on Advances in Financial Technologies},
pages = {29–42},
numpages = {14},
location = {Arlington, Virginia},
series = {AFT '21}
}

@incollection{lamport2019byzantine,
  title={The {Byzantine} generals problem},
  author={Lamport, Leslie and Shostak, Robert and Pease, Marshall},
  booktitle={Concurrency: the works of {Leslie} {Lamport}},
  pages={203--226},
  year={2019}
}

@article{WenLiao2024,
  title={Blockchain service platform evaluation with probabilistic linguistic preference information based on {Choquet} integral and {ExpTODIM}},
  author={Wen, Zhi and Liao, Huchang},
  journal={Information Sciences},
  volume={668},
  pages={120528},
  year={2024},
  publisher={Elsevier}
}

@article{dietvorst2018overcoming,
  title={Overcoming algorithm aversion: People will use imperfect algorithms if they can (even slightly) modify them},
  author={Dietvorst, Berkeley J and Simmons, Joseph P and Massey, Cade},
  journal={Management Science},
  volume={64},
  number={3},
  pages={1155--1170},
  year={2018},
  publisher={INFORMS}
}

@article{GuerreiroEtAl2021,
  title={The hypervolume indicator: Computational problems and algorithms},
  author={Guerreiro, Andreia P and Fonseca, Carlos M and Paquete, Lu{\'\i}s},
  journal={ACM Computing Surveys (CSUR)},
  volume={54},
  number={6},
  pages={1--42},
  year={2021},
  publisher={ACM New York, NY, USA}
}

@article{JacquetLagreze1982,
  title = {Assessing a set of additive utility functions for multicriteria decision-making,  the UTA method},
  volume = {10},
  ISSN = {0377-2217},
  url = {http://dx.doi.org/10.1016/0377-2217(82)90155-2},
  DOI = {10.1016/0377-2217(82)90155-2},
  number = {2},
  journal = {European Journal of Operational Research},
  publisher = {Elsevier BV},
  author = {Jacquet-Lagreze,  E. and Siskos,  J.},
  year = {1982},
  month = June,
  pages = {151–164}
}

@book{Miettinen1999,
  author    = {Miettinen, Kaisa},
  title     = {{Nonlinear Multiobjective Optimization}},
  publisher = {Kluwer Academic Publishers},
  address   = {Boston},
  series    = {International Series in Operations Research \& Management Science},
  volume    = {12},
  year      = {1999},
  doi       = {10.1007/978-1-4615-5563-6}
}

@incollection{MiettinenEtAl2008,
  author    = {Miettinen, Kaisa and Ruiz, Francisco and Wierzbicki, Andrzej P.},
  title     = {{Introduction to Multiobjective Optimization: Interactive Approaches}},
  editor    = {Branke, J{\"u}rgen and Deb, Kalyanmoy and Miettinen, Kaisa and S{\l}owi{\'n}ski, Roman},
  booktitle = {Multiobjective Optimization: Interactive and Evolutionary Approaches},
  series    = {Lecture Notes in Computer Science},
  volume    = {5252},
  pages     = {27--57},
  publisher = {Springer},
  address   = {Berlin, Heidelberg},
  year      = {2008},
  doi       = {10.1007/978-3-540-88908-3_2}
}

@article{XinEtAl2018,
  author  = {Xin, Bin and Chen, Long and Chen, Jie and Ishibuchi, Hisao and Hirota, Kaoru and Liu, Bo},
  title   = {{Interactive Multiobjective Optimization: A Review of the State-of-the-Art}},
  journal = {IEEE Access},
  volume  = {6},
  pages   = {41256--41279},
  year    = {2018},
  doi     = {10.1109/ACCESS.2018.2856832}
}

@incollection{Wierzbicki1980,
  author    = {Wierzbicki, Andrzej P.},
  title     = {{The Use of Reference Objectives in Multiobjective Optimization}},
  editor    = {Fandel, G{\"u}nter and Gal, Tomas},
  booktitle = {Multiple Criteria Decision Making Theory and Application},
  pages     = {468--486},
  publisher = {Springer},
  address   = {Berlin, Heidelberg},
  year      = {1980},
  doi       = {10.1007/978-3-642-48782-8_32}
}

@article{BenayounEtAl1971,
  author  = {Benayoun, Robert and de Montgolfier, Jean and Tergny, Jacques and Laritchev, Oleg},
  title   = {{Linear programming with multiple objective functions: step method ({STEM})}},
  journal = {Mathematical Programming},
  volume  = {1},
  number  = {1},
  pages   = {366--375},
  year    = {1971},
  doi     = {10.1007/BF01584098}
}

@article{ZiontsWallenius1976,
  author  = {Zionts, Stanley and Wallenius, Jyrki},
  title   = {{An Interactive Programming Method for Solving the Multiple Criteria Problem}},
  journal = {Management Science},
  volume  = {22},
  number  = {6},
  pages   = {652--663},
  year    = {1976},
  doi     = {10.1287/mnsc.22.6.652}
}

@article{MiettinenMakela1995,
  author  = {Miettinen, Kaisa and M{\"a}kel{\"a}, Marko M.},
  title   = {{Interactive Bundle-Based Method for Nondifferentiable Multiobjective Optimization: {NIMBUS}}},
  journal = {Optimization},
  volume  = {34},
  number  = {3},
  pages   = {231--246},
  year    = {1995},
  doi     = {10.1080/02331939508844109}
}

@article{MiettinenEtAl2010,
  author  = {Miettinen, Kaisa and Eskelinen, Petri and Ruiz, Francisco and Luque, Mariano},
  title   = {{NAUTILUS} Method: An Interactive Technique in Multiobjective Optimization Based on the Nadir Point},
  journal = {European Journal of Operational Research},
  volume  = {206},
  number  = {2},
  pages   = {426--434},
  year    = {2010},
  doi     = {10.1016/j.ejor.2010.02.041}
}

@incollection{GrecoEtAl2008,
  author    = {Greco, Salvatore and Matarazzo, Benedetto and S{\l}owi{\'n}ski, Roman},
  title     = {{Dominance-Based Rough Set Approach to Interactive Multiobjective Optimization}},
  editor    = {Branke, J{\"u}rgen and Deb, Kalyanmoy and Miettinen, Kaisa and S{\l}owi{\'n}ski, Roman},
  booktitle = {Multiobjective Optimization: Interactive and Evolutionary Approaches},
  series    = {Lecture Notes in Computer Science},
  volume    = {5252},
  pages     = {121--155},
  publisher = {Springer},
  address   = {Berlin, Heidelberg},
  year      = {2008},
  doi       = {10.1007/978-3-540-88908-3_5}
}

@inproceedings{FuernkranzHuellermeier2016PreferenceLearningRanking,
  author    = {F{\"u}rnkranz, Johannes and H{\"u}llermeier, Eyke},
  title     = {Preference Learning and Ranking},
  booktitle = {Proceedings of the Twenty-Fifth International Joint Conference on Artificial Intelligence},
  pages     = {4209--4213},
  year      = {2016}
}

@article{JarvelinKekalainen2002,
  author  = {J{\"a}rvelin, Kalervo and Kek{\"a}l{\"a}inen, Jaana},
  title   = {Cumulated Gain-Based Evaluation of {IR} Techniques},
  journal = {ACM Transactions on Information Systems},
  volume  = {20},
  number  = {4},
  pages   = {422--446},
  year    = {2002},
  doi     = {10.1145/582415.582418}
}

@article{Pasandideh2015,
  title = {Bi-objective optimization of a multi-product multi-period three-echelon supply chain problem under uncertain environments: NSGA-II and NRGA},
  volume = {292},
  ISSN = {0020-0255},
  url = {http://dx.doi.org/10.1016/j.ins.2014.08.068},
  DOI = {10.1016/j.ins.2014.08.068},
  journal = {Information Sciences},
  publisher = {Elsevier BV},
  author = {Pasandideh,  Seyed Hamid Reza and Niaki,  Seyed Taghi Akhavan and Asadi,  Kobra},
  year = {2015},
  month = Jan,
  pages = {57–74}
}

@article{kumar2026blockchain,
  title={A blockchain based authentication protocol for secure autonomous vehicles communication in acoustic networks},
  author={Kumar, Neeraj and Ali, Rifaqat},
  journal={Information Sciences},
  pages={123361},
  year={2026},
  publisher={Elsevier}
}

@article{sadeghi2024synergy,
  title={Synergy between blockchain technology and internet of medical things in healthcare: A way to sustainable society},
  author={Sadeghi, Mahsa and Mahmoudi, Amin},
  journal={Information Sciences},
  volume={660},
  pages={120049},
  year={2024},
  publisher={Elsevier}
}

\end{document}